\documentclass[11pt]{article}

\usepackage[final]{acl}
\usepackage{booktabs}
\usepackage{times}
\usepackage{url}
\usepackage{array}
\usepackage{latexsym}
\usepackage{multirow} 
\usepackage[utf8]{inputenc} 
\DeclareUnicodeCharacter{2248}{\ensuremath{\approx}}
\DeclareUnicodeCharacter{2212}{\ensuremath{-}}
\usepackage[T1]{fontenc}

\usepackage[utf8]{inputenc}

\usepackage{microtype}

\usepackage{inconsolata}

\usepackage{graphicx}

\title{STQA: A Benchmark for Stock-Focused Tabular Question Answering over Historical and Forecasted Data}
\author{
  \textbf{Baoxu An\textsuperscript{1,2}},
  \textbf{Wenmian Yang\textsuperscript{2,\textdagger}},
  \textbf{Zhensheng Wang\textsuperscript{1,2}},
  \textbf{Weijia Jia\textsuperscript{2,3,\textdagger}}
  \\[0.5em]
  \textsuperscript{1}School of Artificial Intelligence,
  Beijing Normal University, Beijing, PR China
  \\
  \textsuperscript{2}Institute of Artificial Intelligence and Future Networks,
  Beijing Normal University, Zhuhai, PR China
  \\
  \textsuperscript{3}Beijing Normal-Hong Kong Baptist University,
  Zhuhai, PR China
  \\[0.5em]
  {\footnotesize
    \{anbaoxu, jensenwang\}@mail.bnu.edu.cn, 
    \{wenmianyang, jiawj\}@bnu.edu.cn
  }
}

\begin{document}
\maketitle
\begin{abstract}
Stock market analysis inherently requires composite reasoning over historical records and future projections, yet existing benchmarks remain fragmented across isolated tasks. We introduce STQA (Stock-focused Tabular Question Answering), an end-to-end benchmark designed to systematically evaluate natural-language question answering over historical data, numerical forecasts, and forecast-based reasoning. Built on a large-scale financial dataset, STQA covers 4,417 stocks and contains 31,400 question–answer pairs derived from expert-crafted templates, accompanied by fine-grained intent and slot annotations. To operationalize this benchmark, we present SQFRS (Stock Query–Forecast–Reasoning System), an agent-based unified framework that orchestrates SQL retrieval and time-series forecasting tools. Experiments demonstrate that while current large language models perform well on historical queries, forecast-based reasoning poses a substantial challenge, revealing critical bottlenecks in tool coordination and reasoning under uncertainty. The dataset and code are available at
\url{https://github.com/xuxubaobaoan/STQA_Project}. 
\end{abstract}



\section{Introduction}
Stock market analysis is driven by structured, time-stamped numerical data such as daily open, close, high, and low prices, trading volumes, and technical indicators. Around this data, three technical paradigms have emerged: structured data retrieval, including semantic parsing and Text-to-SQL interfaces for querying stock databases \cite{pourreza2023dinsqldecomposedincontextlearning}; time-series forecasting, from classical ARIMA models to Transformer-based architectures \cite{3295222.3295349,10.1145/3711896.3737157}; and large language model (LLM)–based financial analysis for macroeconomic understanding and financial document reasoning \cite{nie2024surveylargelanguagemodels,hussain2025artemisdaadvancedreasoningtransformation,10818583,zhang2024multimodalfoundationagentfinancial}.

These paradigms, however, have largely been studied in isolation. Real-world financial analysis is composite: users pose natural-language questions that implicitly require querying historical tables, invoking forecasting models, and then drawing conclusions based on predicted values. Existing systems typically treat these steps as separate tasks, leaving users to decompose questions, run tools in the right order, and integrate partial results by themselves. This paradigm fragmentation limits the usability of language-based financial assistants and makes it hard to systematically study LLMs that must coordinate multiple tools over structured data.

We consider three representative types of questions in this setting.  
(1) Historical query reasoning (e.g., ``Last Friday, which stock, NVIDIA or AMD, had the highest price increase?'') requires factual retrieval and deterministic computation over historical tables \cite{Wang2025}.  
(2) Numerical forecasting (e.g., ``Please forecast NVIDIA’s closing prices for the next five trading days.'') requires modeling future trends from historical time-series data \cite{10.1145/3711896.3737157}. 
(3) Forecast-based reasoning (e.g., ``Will NVIDIA’s highest closing price next month exceed 1{,}000 USD?'') requires logical judgment and threshold comparison over forecasted quantities \cite{zhu2025dianjinr1evaluatingenhancingfinancial}.  
Among these, forecast-based reasoning is crucial in practice, yet rarely evaluated: most existing benchmarks and systems focus on factual historical reasoning and do not test whether a model can invoke a forecaster, interpret its outputs, and reason over them to answer high-level natural-language questions.

Building an end-to-end system that handles all three question types from natural language raises three core challenges. First, accurate interpretation of complex queries: the system must identify the underlying task (historical query, numerical forecasting, or forecast-based reasoning) and extract arguments such as tickers, time ranges, aggregation operations, comparison operators, and thresholds, mapping diverse questions to precise executable specifications. Second, adaptive planning and execution of analytical workflows: different questions require different toolchains and execution orders, potentially combining SQL queries with time-series forecasting in a single workflow that the system must plan and execute autonomously. Third, faithful, evidence-based answer generation: after invoking tools, the system must generate answers strictly grounded in returned evidence—database records or probabilistic forecasts—rather than hallucinated content. In finance, where errors are costly and conclusions must be verifiable, controlling hallucination and ensuring traceable reasoning is crucial \cite{wang2023donotanswerdatasetevaluatingsafeguards,es-etal-2024-ragas,min-etal-2023-factscore}.

There has been rapid progress on tabular question answering and Text-to-SQL, on time-series forecasting, and on tool-augmented LLM agents. Yet, to our knowledge, there is no benchmark that systematically evaluates natural-language to multi-tool workflows combining historical queries, forecasting, and forecast-based reasoning in a unified setting \cite{yuan2025forecastfutureoutcomereasoning}. Existing tabular QA datasets mainly test factual reasoning over static tables, forecasting benchmarks rarely involve natural-language interfaces, and agent benchmarks seldom require precise coordination of database querying and numerical forecasting. This gap makes it difficult to rigorously study how LLMs can understand complex user questions, decide which tools to call and with what arguments, and reason over the resulting outputs within a single environment.

To address this gap, we contribute on both the data and method fronts, with an emphasis on establishing a benchmark rather than a new algorithmic paradigm. On the data side, we construct and release STQA (Stock-focused Tabular Question Answering), a benchmark for language-based agents in stock analysis. Built upon the large-scale FNSPID financial dataset \cite{10.1145/3637528.3671629}, STQA samples historical tables for 4{,}417 stocks, each with six daily fields: date, opening price, closing price, high price, low price, and trading volume. We design 72 question templates (42 historical query reasoning, 16 numerical forecasting, and 14 forecast-based reasoning), instantiate them with real data, and rewrite them into natural language, yielding 31{,}400 question–answer pairs. The forecast-based reasoning questions require logical inference over forecasted values, going beyond traditional factual QA. We additionally provide semantic annotations—intent labels and slots such as ticker, time range, and operations—offering supervision for mapping natural-language questions to structured analytical specifications.

On the method side, we introduce SQFRS (Stock Query–Forecast–Reasoning System) as a unified baseline framework for STQA. SQFRS adopts an agent-based paradigm with an LLM controller: a natural-language understanding module parses questions into intents and slots; a planner routes each parsed task to a predefined workflow that invokes SQL databases and time-series forecasting models; and an answer generator aggregates tool outputs and performs reasoning strictly based on this evidence. SQFRS is not intended as a novel algorithmic contribution, but as a strong, transparent baseline that operationalizes STQA’s three task types in a single system.

Our main contributions are summarized as follows:
\begin{itemize}
    \item We introduce STQA, a benchmark for end-to-end stock analysis QA that unifies three core task types from natural-language questions, providing data resources and evaluation protocols for composite, tool-grounded QA.
    \item We propose SQFRS, an agent-based unified analysis framework that interprets complex questions and orchestrates multi-step workflows across databases and forecasting models, providing a strong baseline on STQA with evidence-grounded answers.
    \item We conduct extensive experiments on STQA that quantify task difficulty and show that SQFRS improves the performance of existing LLMs on this benchmark.
    
\end{itemize}

\begin{table*}
  \centering
  \small
  \begin{tabular}{lcccccc}
    \hline
    \textbf{Dataset} & \textbf{QA Pairs} & \textbf{Tables} & \textbf{Answer format} & \textbf{Forecast Tasks} & \textbf{Multi-scenario Tasks}  \\ 
    \hline
    FinDER\cite{10.1145/3768292.3770361} & 5,703 & 490 & Text & × & × \\
    FinTextQA\cite{Chen_2024} & 1262 & - & Text & × & × \\
    TQABench\cite{qiu2024tqabenchevaluatingllmsmultitable} & 26,260 & 1,200 & Text/SQL & × & ×  \\
    WikiSQL\cite{zhong2017seq2sqlgeneratingstructuredqueries} & 80,654 & 24,241 & SQL & × & ×  \\
    WikiTableQuestion\cite{kweon-etal-2023-open} & 22,033 & 2,108 & Text & ×  & ×  \\
    Open-WikiTable\cite{kweon-etal-2023-open} & 67,023 & 24,680 & Text/SQL & × & × \\
    \textbf{STQA (Ours)} & \textbf{31,400} & \textbf{4,417} & \textbf{Text/SQL} & \textbf{$\surd$} & \textbf{$\surd$} \\
    \hline
  \end{tabular}
  \caption{Dataset comparison, where “-” indicates data that is currently inaccessible.}
  \label{Table 1}
\end{table*}

\section{Related Work}

\subsection{QA datasets}
Table QA (TQA) ranges from open-domain NL-to-SQL tasks \cite{pasupat-liang-2015-compositional,zhong2017seq2sqlgeneratingstructuredqueries,kweon-etal-2023-open} to financial datasets incorporating numerical reasoning \cite{chen-etal-2021-finqa,Wang2025}. However, these benchmarks generally lack unified support for scenarios combining historical retrieval with future forecasting and rely on limited intent diversity. STQA addresses these limitations by integrating query reasoning, numerical forecasting, and forecast-based reasoning within a single benchmark supported by explicit intent annotations. A comparison with existing datasets is given in Table \ref{Table 1}.


\subsection{QA methods}
Early studies on stock-domain QA focused on historical-record QA or simple trend forecasting \cite{kurisinkel2024text2timeseriesenhancingfinancialforecasting}. Recent methods incorporate pre-trained models and structured reasoning, including FinQANet \cite{chen-etal-2021-finqa}, vertical table-QA systems \cite{Wang2025}, and Time-MQA \cite{kong-etal-2025-time}, yet remain limited in task coverage and cross-scenario generalization. LLM-based agents, especially Planner–Executor–Reasoner (PER) architectures \cite{christakopoulou2024agentsthinkingfastslow,qiao-etal-2024-autoact}, improve tool use and interpretability, but their fixed pipelines hinder complex analytical tasks and composite stock queries requiring flexible coordination of database and forecasting tools \cite{li2025designingdomainspecificagentshierarchical,lin-etal-2025-llm,zeng2025futurexadvancedlivebenchmark,karger2025forecastbenchdynamicbenchmarkai}. To address these limitations, our SQFRS plans multi-step workflows via precise natural-language understanding and dynamically coordinates SQL retrieval with time-series forecasting for complex stock-domain QA.

\section{Dataset Construction and Analysis}

\subsection{Stock Data Collection and Preprocessing}
STQA is built on top of the large-scale public financial dataset FNSPID \cite{10.1145/3637528.3671629}, which contains 4,475 stock-specific tables. From each table, we extract five core daily fields that are most relevant for our downstream tasks: open, close, high, low, and volume.

Because raw financial time series have missing values, irregular trading calendars, and inconsistent lengths, we apply a two-step preprocessing pipeline. First, we perform temporal alignment: we discard tables whose time span does not fully cover 2021–2023 and align all remaining stocks to a shared trading calendar. Second, we normalize trading days and sequence length. Since equity markets are closed on weekends, we remove Saturdays and Sundays and directly concatenate records from Monday to Friday so that each stock is defined only on actual trading days. From each aligned trading-day sequence, we then extract a continuous subsequence of 700 trading days, obtaining a fixed-length representation per stock that downstream models and tools can directly consume.

After preprocessing, 4,417 high-quality tables remain. We split them into training, validation, and test sets with a 7:2:1 ratio, yielding 3,094 training tables, 879 validation tables, and 444 test tables.




\subsection{Question–Answer Generation}
To approximate the complex, multi-dimensional analysis needs of real investors, we design 72 natural language query templates covering three task categories: query reasoning, numerical forecasting, and forecast-based reasoning. These templates are instantiated with concrete stocks, dates, and values from the preprocessed tables to form grounded QA pairs.

Because the underlying data are defined only over trading days, all questions and answers are implicitly workday-based. Weekend dates never appear as query conditions or targets, so market closures do not introduce ambiguity into either the time series or the QA labels.




\subsubsection{Query Reasoning}
Query reasoning is implemented with 42 templates, including 32 single-table templates and 10 multi-table templates. They cover all three numeric fields (opening price, closing price, trading volume), a wide range of temporal expressions (exact dates and vague times such as relative offsets and holiday-based descriptions), and multiple task types (vague-time queries, comparison queries, complex multi-step reasoning, and numerical reasoning). The multi-table templates further support cross-stock and multi-time-point queries and cross-table numerical reasoning over both stock and time dimensions, providing a structured benchmark for composite reasoning on financial time series.

\subsubsection{Numerical Forecasting}
Numeric Forecasting is implemented with 16 templates, including 3 single-time-point templates and 13 vague-time templates. Since short-term trading data are often affected by random fluctuations, long-term trends, and sudden information shocks, we regard trading volume as an unstable forecasting target. Therefore, all templates focus on four core price fields: opening price, closing price, highest price, and lowest price. Built upon the conventional text-to-SQL paradigm, this task further introduces a time-series forecasting step: the system is required not only to parse and align temporal expressions, but also to retrieve the corresponding historical opening, closing, highest, and lowest price sequences from structured tables and subsequently forecast future prices. This task therefore provides a structured benchmark for jointly evaluating natural language understanding, temporal data retrieval, and time-series forecasting capabilities.

\subsubsection{Forecast-based Reasoning}
Forecast-based reasoning is designed with 14 core templates that cover composite tasks ranging from single-point trend forecasting and reasoning, to fuzzy-time forecasting and reasoning, to multi-time-point forecasting and reasoning, and further to more complex logical reasoning. Unlike numerical forecasting tasks that focus only on price prediction, this category further emphasizes logical analysis over the forecast results and primarily centers on two core price fields: Open (opening price) and Close (closing price). Each template integrates three capabilities: text‑to‑SQL style data retrieval, time‑series forecasting, and logical reasoning over the forecast results (e.g., judging trends, records, or threshold conditions over future periods). This category is intended to assess models’ end‑to‑end ability to connect quantitative forecasts with higher‑level, logic‑driven decision queries.



\subsection{Intent and Slot Annotation}
To support precise parsing of user queries, STQA follows task-oriented dialogue practice and adds intent and slot annotations. We define ten intent types reflecting the four numeric fields and typical investor needs: Opening Price Inquiry, Closing Price Inquiry, Volume Inquiry, Opening Price Forecasting, Closing Price Forecasting, Opening Price Trend Forecasting, Closing Price Trend Forecasting, Opening Price Extremum Forecasting, Closing Price Extremum Forecasting, and Daily Extremum Price Forecasting. Together, these intents cover both retrospective queries about historical values and forward-looking queries about future prices, trends, and extrema.

We also define five slot types to capture key entities and temporal attributes: Stock Tickers, Year, Month, Day, and Fuzzy Date. The Fuzzy Date slot encodes relative temporal expressions, which are normalized with respect to the global trading calendar established in preprocessing. Detailed definitions and corpus-level statistics for all intents and slots are provided in appendix \ref{sec:appendixC}.

\subsection{Query Rewriting and Quality Control}
Template-based generation ensures semantic controllability and answer correctness, but often yields rigid and repetitive expressions that deviate from natural user language. To improve linguistic quality, we use GPT-oss-120B to paraphrase the template-generated questions. The prompts require synonym substitution, local rephrasing, and syntactic restructuring while strictly preserving the original meaning; entities, dates, and numerical conditions must remain unchanged.

All paraphrased QA pairs are then manually reviewed. Annotators check grammaticality, naturalness, and exact semantic equivalence to the original template instances, and discard or correct cases with meaning drift, ambiguity, or inconsistency with the underlying tables.

After rewriting and quality control, STQA contains 31,400 high-quality QA pairs. To quantify linguistic improvement, we use Qwen3-30B, which is different from the rewriter, to score query naturalness on a 0--4 scale, where 0 denotes rigid template style and 4 denotes natural human writing. The average score increases from 3.18 for the original templates to 3.36 for the final corpus, indicating improved diversity and fluency without sacrificing semantic fidelity (see Appendix \ref{sec:appendixE}). We further conduct a human evaluation with eight volunteers, who assess 50 randomly sampled test questions using the same 0--4 scale. The average score rises from 2.48 before rewriting to 2.84 after rewriting, further confirming the gains observed in automatic evaluation (see Appendix \ref{sec:appendixE}).

The 31,400 instances are split into training, validation, and test sets at a 7:2:1 ratio, yielding 21,980, 6,280, and 3,140 instances, respectively. By task type, STQA includes 10,500 query reasoning QA pairs, 10,400 numerical forecasting QA pairs, and 10,500 forecast-based reasoning QA pairs. Additional descriptive statistics and examples are provided in Appendix \ref{sec:appendixC}.


STQA adopts a unified hierarchical labeling scheme across all question types, where each query is annotated with its intent, slots, and ground-truth answer. For Query Reasoning questions, we additionally provide the gold SQL statement that returns the correct answer from the stock tables. For Numerical Forecasting and Forecast-based Reasoning questions, we provide the historical time-series data required for reasoning, along with the SQL queries used to retrieve it. Detailed dataset statistics and the full annotation schema are reported in Appendix \ref{sec:appendixC}.

\section{Method}
\subsection{Overview}
\begin{figure*}[t]
  \centering
  \includegraphics[width=0.96\textwidth]{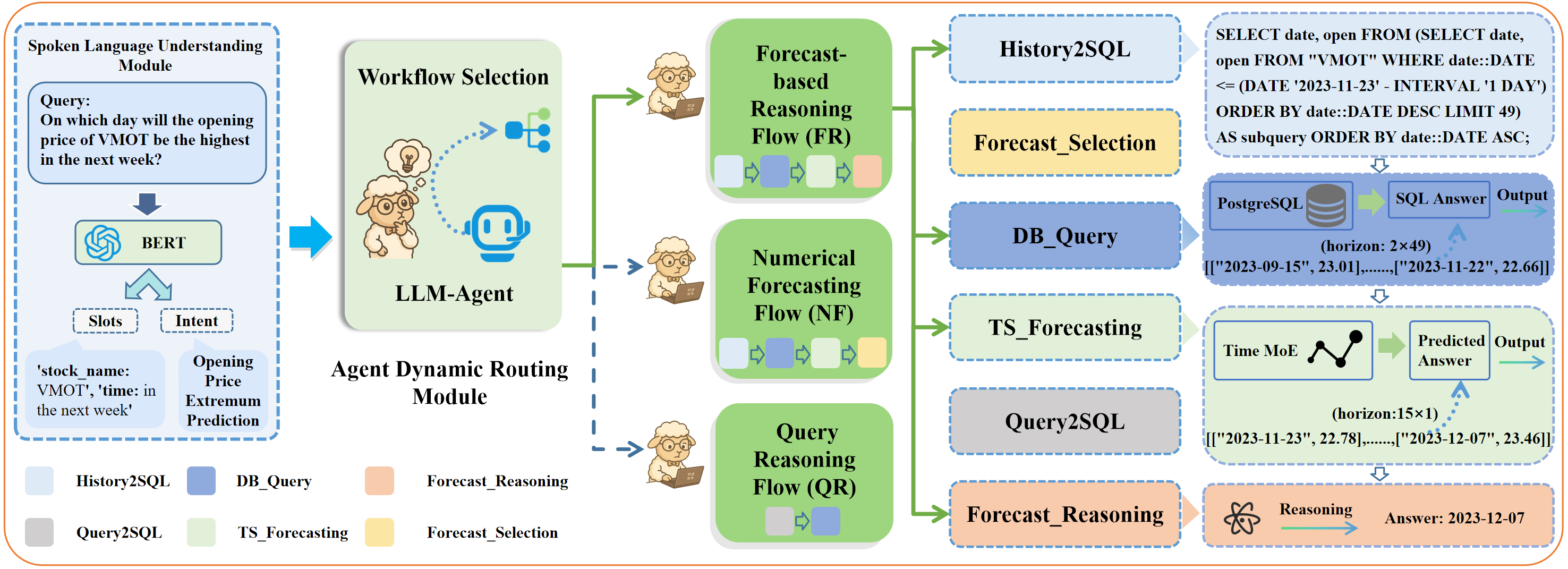}
  \caption{The overall framework of SQFRS. Given an input query, the SLU module predicts its intent and extracts slots. In the illustrated example, the Dynamic Routing Agent identifies the query as a forecast-based reasoning problem and selects the corresponding FR workflow to generate the final answer.}
  \label{fig:experiments1}
  \vspace{-1em}
\end{figure*}

To address complex stock analysis scenarios, we propose SQFRS (Stock Query – Forecast – Reasoning System), an agent-based unified framework that serves as a strong baseline on STQA. SQFRS uses an LLM-based dynamic routing agent as the central controller to coordinate specialized modules for query reasoning, numerical forecasting, and forecast-based reasoning within a unified architecture. The overall framework is shown in Figure \ref{fig:experiments1}.

Given a user query, a Spoken Language Understanding (SLU) module first predicts intent and slots, which guide the routing agent to select one of three pre-defined workflows and dispatch the corresponding tools. This agent-centered ``understand–route–execute'' design enables flexible, intent-aware tool composition with controlled logic and execution efficiency. We next describe each module, the agent, and the workflows.

\subsection{SLU Module}
SLU in our system follows the standard task-oriented dialogue setting and includes two types of labels: intents and slots. Intent labels capture the user’s high-level goal, while slot labels extract key arguments such as stock tickers, years, months, and specific or fuzzy dates.

In this paper, we build an SLU module to predict intent and slot labels for each query, and use these structured labels to provide additional guidance to the dynamic routing agent. Following previous work \cite{Wang2025}, we fine-tune a BERT-based model \cite{devlin2019bertpretrainingdeepbidirectional} to implement the SLU module and perform intent detection and slot filling.

\subsection{Dynamic Routing Agent}
The Dynamic Routing Agent (DRA) is the decision core of SQFRS. It consumes three inputs: the intent label, slot values from the SLU module, and the original natural-language query. Based on these signals, it selects one of three workflows: Query Flow, Forecast Flow, or Forecast-Reasoning Flow.

Routing is realized via a structured system prompt, the Agent Dynamic Routing Prompt (see appendix \ref{sec:appendixF}). The prompt (i) specifies system capability boundaries to the LLM, (ii) enumerates the available tools, i.e., \textit{Query2SQL}, \textit{History2SQL}, \textit{Database Query}, \textit{Time Series Forecasting}, \textit{Forecast Selection}, and \textit{Forecast-based Reasoning}, with brief functional descriptions and I/O schemas, and (iii) introduces admissible tool-calling orders for each workflow. Concretely, \textbf{Query Reasoning (QR)} calls \textit{Query2SQL} then \textit{Database Query}; \textbf{Numerical Forecasting (NF)} calls \textit{History2SQL}, \textit{Database Query}, \textit{Time Series Forecasting}, and \textit{Forecast Selection (FS)}; \textbf{Forecast-based Reasoning (FR)} reuses \textit{History2SQL}, \textit{Database Query}, and \textit{Time Series Forecasting}, but replaces \textit{Forecast Selection} with \textit{Forecast-based Reasoning}.

Under these constraints, the DRA maps user intent to a valid workflow and tool chain, then executes the selected workflow by invoking tools step by step according to the pre-defined orchestration logic.

\subsection{SQL Generation Module}
The SQL Generation module converts a natural-language query into an executable SQL statement based on the intent and slot information produced by the SLU module. We employ an LLM and design two submodules, i.e., \textit{Query2SQL} and \textit{History2SQL}, for query workflows and forecasting workflows, respectively. \textit{Query2SQL} generates SQL statements for structured data retrieval, whereas \textit{History2SQL} retrieves historical time-series data to support downstream forecasting and reasoning.

Both submodules follow a unified generation framework, differing only in their functional objectives. For each submodule, we design dedicated prompts consisting of a task description, in-context examples, and the current input (see Appendix \ref{sec:appendixF}). The examples are dynamically retrieved from training instances of the same task type and include the query, slots, and corresponding gold SQL. By exposing semantic correspondences and SQL patterns from similar cases, this design encourages the model to reuse appropriate logical structures.

\subsection{Database Query Module}
The Database Query module is implemented on PostgreSQL, where all stock tables are stored. It takes an SQL statement from Query2SQL or History2SQL, executes it, and returns standardized results.

\subsection{Time Series Forecasting Module}

The Time Series Forecasting module uses the standardized historical data returned by the Database Query module, driven by History2SQL, to predict stock prices for the next 15 trading days. It converts the date--price sequence into a one-dimensional numerical series and feeds it into an externally trained stock forecasting model (see Section~5 for experimental details).

The outputs are then aligned with the trading calendar. Weekend dates within the 15-day horizon are assigned zero prices through post-processing, while real trading days are retained. During training and evaluation, zero-price dates are excluded from metric computation, so the reported scores measure forecasting accuracy on trading days only. This yields an automated pipeline covering historical retrieval, series conversion, model inference, and calendar-aware normalization.

\subsection{Forecast Decision Module}

The Forecast Decision module provides a unified decision-making framework over time-series forecasting results to generate final answers for forecasting-related tasks. It consists of two submodules, i.e., \textit{Forecast Selection} and \textit{Forecast-based Reasoning}, both of which rely on an LLM with task-specific prompts. For both submodules, the prompts are instantiated with in-context examples dynamically retrieved from training instances sharing the same intent as the current query (see Appendix \ref{sec:appendixF}), enhancing the mapping between task semantics and target outputs.

The Forecast Selection module handles tasks whose answers can be directly located in the forecast sequence. It integrates the user query, the intent and slots produced by the SLU module, and the forecast sequence generated by the time-series forecasting module (\texttt{Pred\_answer}). The model is then guided to identify the target position in \texttt{Pred\_answer} using key slot values such as stock ticker and normalized date, and to extract the corresponding forecast value. Each in-context example contains the original query, SLU parse, forecast sequence, and normalized gold answer, showing how structured semantics map to the correct element in the forecast sequence.

The Forecast-based Reasoning module targets judgment-oriented and extreme-value queries that cannot be directly handled by the standard tool chain. It jointly reasons over the natural-language query, the historical series (\texttt{extracted\_history}), and the forecast series (\texttt{Pred\_answer}) to produce the final answer. Unlike Forecast Selection, its in-context examples additionally include a canonical reasoning trajectory, explicitly demonstrating the logical path from input information to the inferred conclusion and guiding the model toward more faithful and stable reasoning.

\section{Experiment}

\begin{table*}[tp]
  \centering
  \small
  \begin{tabular}{lcccccccccccc}
    \hline
    &  & \multicolumn{2}{c}{\textbf{QR}} & \multicolumn{3}{c}{\textbf{NF}} & \multicolumn{2}{c}{\textbf{FR}} & \multicolumn{1}{c}{\textbf{WF}}\\
    \cmidrule(lr){3-4} \cmidrule(lr){5-7} \cmidrule(lr){8-9} \cmidrule(lr){10-10}
    \textbf{Model} & \textbf{Method} & \textbf{F1} & \textbf{Acc} & \textbf{MSE} & \textbf{MAE} & \textbf{MRE} & \textbf{F1} & \textbf{Acc} & \textbf{Acc} \\
    \hline
    \multirow{2}{*}{Qwen3 30B} 
      & SQFRS   & 0.7000 & 0.6924 & 81.3165 & 2.4583 & 0.0617 & 0.5536 & 0.5529 & 99.67\%\\
      & Vanilla & 0.6755 & 0.6406 & 128.6482 & 2.8026 & 0.0744 & 0.5111 & 0.4672 & -\\
    \hline
    \multirow{2}{*}{GPT-oss-20B} 
      & SQFRS   & 0.7186 & 0.7152 & 81.3120 & 2.4561 & 0.0616 & 0.6234 & 0.5929 & 99.93\%\\
      & Vanilla & 0.6259 & 0.6174 & 126.2028 & 2.7814 & 0.0742 & 0.5202 & 0.4706 & -\\
    \hline
    \multirow{2}{*}{GPT-oss-120B} 
      & SQFRS   & 0.7309 & 0.7178 & \textbf{80.5744} & \textbf{2.4140} & \textbf{0.0590} & 0.6319 & 0.6171 & 100\%\\
      & Vanilla & 0.6328 & 0.6190 & 117.7436 & 2.7461 & 0.0716 & 0.5717 & 0.4848 & -\\
    \hline
    \multirow{2}{*}{GEMMA 4-31B} 
      & SQFRS   & \textbf{0.7370} & \textbf{0.7238} & 81.4815 & 2.4828 & 0.0623 & \textbf{0.6798} & \textbf{0.6243} & \textbf{100}\%\\
      & Vanilla & 0.6308 & 0.6185 & 124.8661 & 2.7677 & 0.0727 & 0.5250 & 0.4743  & -\\
    \hline
    
  \end{tabular}
  \caption{\label{Table 2}
   Overall performance comparison between the SQFRS framework and the Vanilla baseline, with boldface indicating the best result achieved by each method. “–” indicates that WF is not applicable.}
   \vspace{-0.5em}
\end{table*}

\subsection{Evaluation Metrics and Baselines}

\textbf{Evaluation Metrics.} Reasoning tasks (Query Reasoning and Forecast-based Reasoning) are evaluated using Accuracy (Acc) and F1-score (F1). Acc is a strict exact-match metric requiring all predicted items to match the ground truth, while F1 computes the harmonic mean of precision and recall at the item level, granting partial credit for multi-item answers. For numeric forecasting tasks, we use Mean Absolute Error (MAE), Mean Squared Error (MSE), and Mean Relative Error (MRE).

\textbf{Baselines and Settings.} In the SQFRS framework, the LLM agents are instantiated with four open-source LLMs: Qwen3-30B \cite{yang2025qwen3technicalreport}, GPT-oss-20B/120B \cite{openai2025gptoss120bgptoss20bmodel}, and Gemma 4-31B \cite{google2026gemma4modelcard}.

To identify the optimal forecaster for Time Series Forecasting Module, we evaluate six models: the general-purpose LLM GPT-oss-120B, the state-of-the-art LLM-based model Time-MoE \cite{shi2025timemoebillionscaletimeseries}, and four lightweight models: TimesNet \cite{wu2023timesnettemporal2dvariationmodeling}, I-Transformer \cite{liu2024itransformerinvertedtransformerseffective}, TimeMixer \cite{ICLR2024_a7ac8a21}, and TimeXer \cite{NEURIPS2024_0113ef46}. Based on evaluation results across multiple fields, including open, close, high, and low, Time-MoE achieves the best average performance and is therefore selected as the default forecaster (see Appendix \ref{sec:appendixB} for details). All time-series-related tasks, including Numerical Forecasting and Forecast-based Reasoning, use the same forecasting module across LLM backbones. Thus, the forecasting model is strictly controlled, and performance differences mainly reflect each LLM's ability to retrieve historical data, extract forecast values, and perform forecast-based reasoning.

The SLU module is fine-tuned from a pre-trained BERT model\footnote{\url{https://huggingface.co/google-bert/bert-base-cased}} on the training set for 10 epochs.

As STQA is a newly proposed benchmark, we introduce a \textit{Vanilla} baseline, which disables the workflow mechanism, allowing the Dynamic Routing Agent (DRA) to invoke modules autonomously.




\subsection{Results and Analysis}

As shown in Table~\ref{Table 2}, SQFRS consistently outperforms the Vanilla baseline, where the workflow mechanism is disabled and the DRA invokes modules autonomously, across all evaluated backbone models. On average, removing the workflow leads to a 8.8425\% accuracy drop on Query Reasoning (QR) tasks and a 12.2575\%  accuracy drop on Forecast-based Reasoning (FR) tasks, together with a 0.01202 increase in MRE on Numeric Forecasting (NF) tasks. Meanwhile, the DRA in SQFRS achieves a routing accuracy exceeding 99.9\% on average, indicating that task partitioning errors are negligible. These results demonstrate that the workflow mechanism provides reliable task routing and improves the accuracy of both direct reasoning and forecast-based decision making.

Across backbone models, GEMMA 4-31B achieves the best overall performance on reasoning-oriented tasks, with 72.38\% accuracy on QR and 62.43\% accuracy on FR, whereas Qwen3-30B performs the worst on these two tasks. For NF, GPT-oss-120B obtains the lowest MRE of 0.0590, while GEMMA 4-31B yields the worst MRE (see Table \ref{Table 7} for detailed results). At the task level, SQFRS shows a clear performance alignment between QR and FR: models with stronger QR performance generally achieve better FR results. However, lower MRE on NF does not necessarily translate into higher FR accuracy, suggesting that numerical forecasting mainly measures value-level closeness, whereas FR further depends on whether the predicted values support the correct reasoning judgment. The role of historical retrieval quality in these results is further discussed in Appendix \ref{sec:appendixA.2}.

\begin{table}
  \centering
  \small
  \begin{tabular}{lcccc}
    \hline
    \multicolumn{1}{c}{\multirow{2}{*}{\textbf{Model}}} & 
    {\multirow{2}{*}{\textbf{Method}}} & 
    \textbf{QR} & \textbf{NF} & \textbf{FR}\\
    \cmidrule(lr){3-3} \cmidrule(lr){4-4} \cmidrule(lr){5-5}
    & & \textbf{Acc} & \textbf{MRE} & \textbf{Acc} \\
    \hline
    \multirow{2}{*}{Qwen3 30B} 
      & SQFRS   & 0.6924 & 0.0617 & 0.5529 \\
      & W/O SLU & 0.6486 & 0.0704 & 0.5029 \\
    \hline
    \multirow{2}{*}{GPT-oss-20B} 
      & SQFRS   & 0.7152 & 0.0616 & 0.5929 \\
      & W/O SLU & 0.6705 & 0.0704 & 0.5509 \\
    \hline
    \multirow{2}{*}{GPT-oss-120B} 
      & SQFRS   & 0.7178 & 0.0590 & 0.6171\\
      & W/O SLU & 0.6738 & 0.0700 & 0.5671 \\
    \hline
    \multirow{2}{*}{GEMMA 4-31B}  
      & SQFRS   & 0.7238 & 0.0623 & 0.6243 \\
      & W/O SLU & 0.6532 & 0.0708 & 0.5700 \\
    \hline
        \multirow{2}{*}{Average} 
      & SQFRS   & 0.7123 & 0.0612 & 0.5968 \\
      & W/O SLU & 0.6615 & 0.0704 & 0.5477 \\
    \hline
  \end{tabular}
  \caption{\label{Table 3}
    Ablation Study Results for the SLU Module.}
\vspace{-1em}
\end{table}

\subsection{Ablation Study}

\begin{figure}[t]
\centering
  \includegraphics[width=0.85\linewidth]{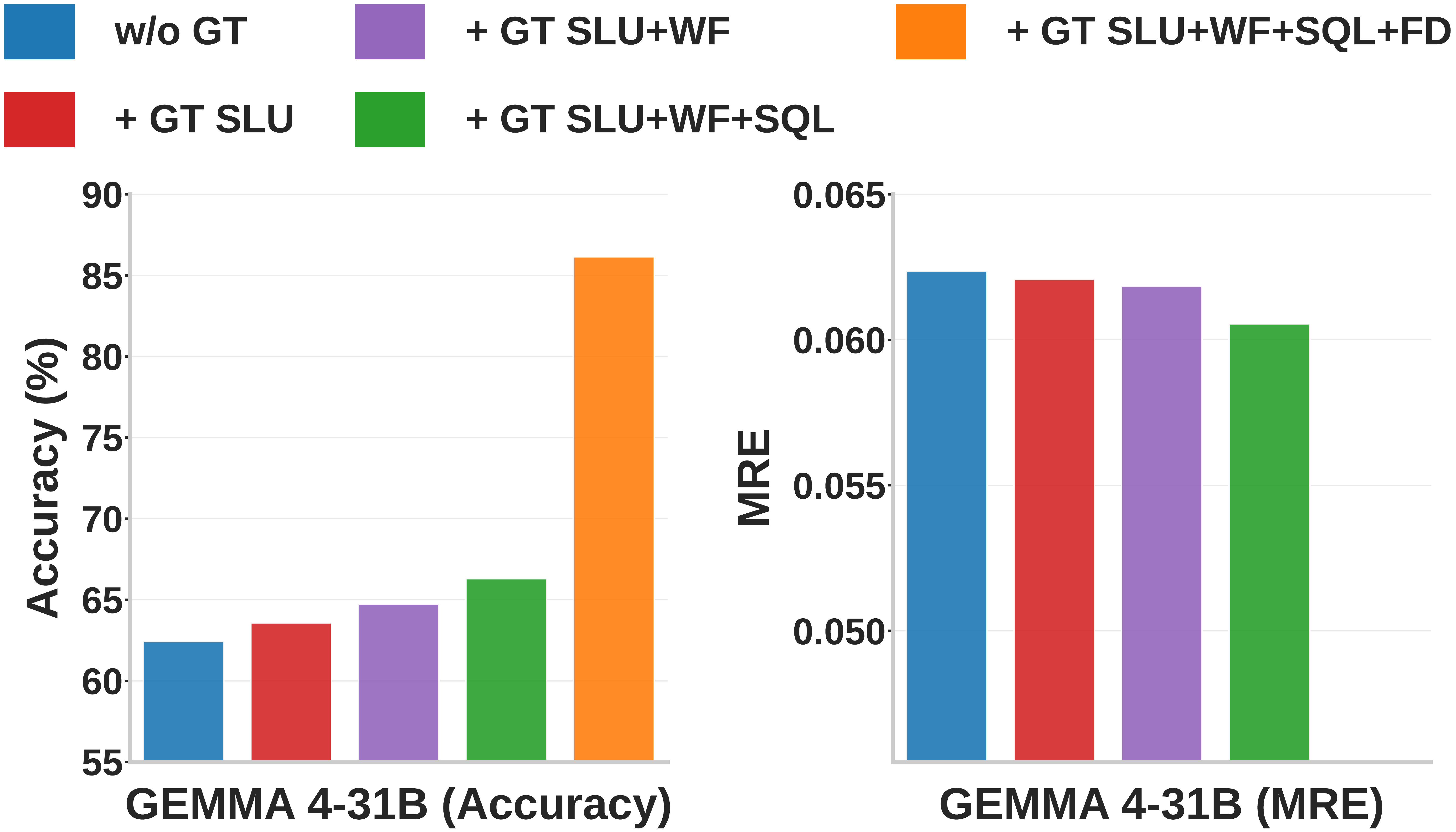}
  \caption {The impact of different GT labels.}
  \label{fig:GEMMA_4-31B_Taller_Bars}
  \vspace{-1em}
\end{figure}

We conduct ablation studies to examine the contributions and bottlenecks of SQFRS. First, we evaluate the effect of the SLU module by disabling it as a baseline (W/O SLU), as shown in Table~\ref{Table 3}. Compared with W/O SLU (QR: 66.15\%, NF MRE: 0.0704, FR: 54.77\% on average), SQFRS achieves consistent improvements across all tasks (QR: 71.23\%, NF MRE: 0.0612, FR: 59.68\% on average). These gains show that intent labels help the DRA allocate workflows, while slot labels provide constraints for SQL generation, confirming the effectiveness of the SLU module. We further evaluate an LLM-based in-context learning (ICL) alternative for SLU prediction under limited-label settings in Appendix~\ref{sec:appendixA.4}. Although it slightly underperforms the fine-tuned BERT SLU module, it substantially outperforms the setting without SLU labels, confirming the necessity and robustness of SLU information.

Second, due to the serial structure of NF and FR workflows, errors may accumulate across components. We therefore conduct a bottleneck analysis by incrementally injecting ground-truth (GT) labels into four key intermediate steps in execution order: SLU labels, workflow routing (WF), historical SQL generation (SQL), and forecast data (FD), to identify the dominant source of error propagation.

We use the GEMMA 4-31B model as a representative example for analysis, shown in Figure \ref{fig:GEMMA_4-31B_Taller_Bars}. Results reveal that GT SLU yields only modest gains (QR: +1.54\%; NF: -0.00028; FR: +1.14\%), indicating that the BERT-based SLU module is relatively robust (see Appendix \ref{sec:appendixA.4}). GT WF provides the smallest improvements for QR and FR (QR: +0.24\%; FR: +0.17\%) and the smallest NF MRE reduction (-0.00023), consistent with the DRA’s high accuracy (99.9\%). GT SQL produces a more pronounced FR gain (+2.55\%) and a larger NF MRE reduction (-0.0013), suggesting that SQL generation remains a bottleneck, while GT FD yields the largest gain for FR (+19.85\%), highlighting significant improvement potential in the time-series forecasting module. Note that the resulting MRE in this setting is too small (0.004931) to visualize clearly in the experimental figure.

Even with GT labels for all intermediate steps, FR accuracy reaches only 86.14\%, indicating that logical reasoning remains a primary bottleneck. Similar patterns across all LLM backbones (see Appendix~\ref{sec:appendixD}) further confirm this finding. We also include representative failure cases in Appendix~\ref{sec:appendixG} to illustrate typical error sources.

\begin{table*}[h]
  \centering
  \small
  \begin{tabular}{lcccccccc}
    \hline
    &  & \multicolumn{2}{c}{\textbf{QR}} & \multicolumn{3}{c}{\textbf{NF}} & \multicolumn{2}{c}{\textbf{FR}}\\
    \cmidrule(lr){3-4} \cmidrule(lr){5-7} \cmidrule(lr){8-9}
    \textbf{Model} & \textbf{Method} & \textbf{F1} & \textbf{Acc} & \textbf{MAE} & \textbf{MRE} & \textbf{MSE} & \textbf{F1} & \textbf{Acc} \\
    \hline
    \multirow{2}{*}{Qwen3 30B} 
      & Template & 0.7000 & 0.6924 & 2.4583 & 0.0617 & 81.3165 & 0.5536 & 0.5529 \\
      & Human    & 0.5303 & 0.5000 & 2.7960 & 0.0743 & 127.5079 & 0.4857 & 0.4750 \\
    \hline
    \multirow{2}{*}{GPT-oss-20B} 
      & Template & 0.7186 & 0.7152 & 2.4561 & 0.0616 & 81.3120 & 0.6234 & 0.5929 \\
      & Human    & 0.5962 & 0.5333 & 2.7714 & 0.0731 & 125.5648 & 0.4970 & 0.4667 \\
    \hline
    \multirow{2}{*}{GPT-oss-120B} 
      & Template & 0.7309 & 0.7178 & 2.4140 & 0.0590 & 80.5744 & 0.6319 & 0.6171 \\
      & Human    & 0.5970 & 0.5667 & 2.7560 & 0.0723 & 121.6482 & 0.5369 & 0.5250 \\
    \hline
    \multirow{2}{*}{GEMMA 4-31B} 
      & Template & 0.7370 & 0.7238 & 2.4828 & 0.0623 & 81.4815 & 0.6798 & 0.6243 \\
      & Human    & 0.6296 & 0.5667 & 2.8296 & 0.0755 & 130.2493 & 0.5000 & 0.5000 \\
    \hline
  \end{tabular}
  \caption{\label{tab:4}
   Result of Real-world Questions}
\end{table*}

To assess the gap between templated and real-world queries, we recruit volunteers to collect 100 human-written questions, including 30 Query Reasoning (QR), 30 Numeric Forecasting (NF), and 40 Forecast-based Reasoning (FR) questions. We then evaluate SQFRS with Qwen3-30B, GPT-oss-20B, GPT-oss-120B, and GEMMA 4-31B on this test set, with detailed results shown in Table~\ref{tab:4}.

The results show that all models performed worse on manually written questions than on template-generated questions, indicating that irregular expressions, colloquial phrasing, and open-ended semantics in real-world queries increase task difficulty. Although manually written questions are more challenging, the results still exhibit performance trends consistent with those for template-generated questions. Specifically, QR performance decreased by an average of 17.06\%, while FR performance decreased by an average of 10.51\%.

\begin{table}[h]
  \centering
  \small
  \begin{tabular}{lcc}
    \hline
    \textbf{Model} & \textbf{Setting} & \textbf{FR Acc (\%)} \\
    \hline
    Human        & Stockholder     & 69.17 \\
    Human        & Non-stockholder & 63.33 \\
    Qwen3 30B    & SQFRS           & 59.17 \\
    Qwen3 30B    & DR              & 54.17 \\
    GPT-oss-20B  & SQFRS           & 62.50 \\
    GPT-oss-20B  & DR              & 50.88 \\
    GPT-oss-120B & SQFRS           & 65.83 \\
    GPT-oss-120B & DR              & 55.83 \\
    GEMMA 4-31B  & SQFRS           & 68.33 \\
    GEMMA 4-31B  & DR              & 56.67 \\
    \hline
  \end{tabular}
  \caption{\label{tab:5} FR Accuracy of Humans and Models under SQFRS vs DR.}
\end{table}

Moreover, we construct a 120-question human-comparison set to compare direct LLM reasoning (DR), SQFRS, and human judgments from volunteers with and without stock-investment experience. We recruit six volunteers for the human evaluation, including four stock investors and two non-investors. The annotators are provided only with the question text and the corresponding ground-truth historical data, without access to any future information. Under the DR setting, the agent can make decisions only based on the historical data returned by History2SQL and is not allowed to perform numerical forecasting or invoke any time-series model.

As shown in Table~\ref{tab:5}, stock investors and non-investors achieve FR accuracies of 69.17\% and 63.33\%, respectively. Under the DR setting, Qwen3-30B, GPT-oss-20B, GPT-oss-120B, and GEMMA 4-31B achieve FR accuracies of 54.17\%, 50.88\%, 55.83\%, and 56.67\%, respectively, remaining close to random-chance performance. By contrast, their corresponding accuracies under the SQFRS setting increase to 59.17\%, 62.50\%, 65.83\%, and 68.33\%. In particular, SQFRS with GEMMA 4-31B achieves performance comparable to that of stock investors, slightly outperforms non-investors, and surpasses all DR baselines. These results demonstrate that the complete toolchain improves performance on FR tasks.

\section{Conclusion}
In this paper, we introduce STQA, a large-scale benchmark for financial tabular question answering that aims to advance end-to-end research integrating multiple paradigms. Built upon real-world stock data, the benchmark comprises 31,400 question-answer pairs grounded in 4,417 stock tables, and unifies three core tasks, i.e., query reasoning, numerical forecasting, and forecast-based reasoning. Additionally, we propose SQFRS as a strong baseline approach. Experimental results not only validate the effectiveness of this method but also underscore the difficulty of the benchmarks.

\section*{Limitation}
This study has several notable limitations. First, while the STQA benchmark primarily utilizes stock market data, this specific focus enables us to evaluate model capabilities in a well-defined context. The measuring methodologies are applicable to various financial tasks, such as investment strategy or credit risk assessment, where similar reasoning patterns are employed.

Second, the models selected for evaluation were chosen based on their relevance and ability to serve as a reliable baseline. This approach is common in the field, focusing on representative models rather than exhaustively testing every available architecture; thus, it provides targeted analysis of the most pertinent models. Future research can expand on this by including additional models to test their performance within this framework.

Additionally, the main contribution of this work lies in constructing the STQA benchmark rather than improving a specific method. Such analyses typically pertain more to method-oriented research rather than framework evaluation; therefore, our focus remains on demonstrating the utility of STQA as an evaluation benchmark and SQFRS as a baseline framework.

Moreover, although the dataset captures a variety of question types, it may not fully represent the complexities of all real-world inquiries. However, it serves its purpose well for evaluating the framework. The dataset was generated using templates, a common practice in large-scale question-answering benchmarks. This allows for systematic variation and consistency in the questions, making it easier to evaluate model performance under controlled conditions. To enhance the diversity of expressions, we employed large language models for rewriting and included assessments from both model evaluations and human sampling, demonstrating that the rewritten expressions achieve a level of richness beyond mere template-based presentation.

In summary, while these limitations exist, they do not detract from the significant contributions made by the study. By articulating these constraints clearly, we aim to provide a foundation for future research that can build upon these insights.

\section*{Acknowledgments}

This work is supported in part by the National Natural Science Foundation of China (NSFC) under Grant No.~62272050 and the grant of Beijing Normal--Hong Kong Baptist University sponsored by the Guangdong Provincial Department of Education; in part by the Zhuhai Science-Tech Innovation Bureau under Grant No.~2320004002772; and by the Interdisciplinary Intelligence Super Computer Center of Beijing Normal University (Zhuhai).

\bibliography{latex/custom}

@inproceedings{pourreza2023dinsqldecomposedincontextlearning,
      title={Din-sql: Decomposed in-context learning of text-to-sql with self-correction},
  author={Pourreza, Mohammadreza and Rafiei, Davood},
  booktitle={Advances in neural information processing systems},
  volume={36},
  pages={36339--36348},
  year={2023}
}

@inproceedings{3295222.3295349,
author = {Vaswani, Ashish and Shazeer, Noam and Parmar, Niki and Uszkoreit, Jakob and Jones, Llion and Gomez, Aidan N. and Kaiser, {\L}ukasz and Polosukhin, Illia},
title = {Attention is all you need},
year = {2017},
isbn = {9781510860964},
publisher = {Curran Associates Inc.},
address = {Red Hook, NY, USA},
booktitle = {Proceedings of the 31st International Conference on Neural Information Processing Systems},
pages = {6000–6010},
numpages = {11},
location = {Long Beach, California, USA},
series = {NIPS'17}
}

@misc{nie2024surveylargelanguagemodels,
      title={A Survey of Large Language Models for Financial Applications: Progress, Prospects and Challenges}, 
      author={Yuqi Nie and Yaxuan Kong and Xiaowen Dong and John M. Mulvey and H. Vincent Poor and Qingsong Wen and Stefan Zohren},
      year={2024},
      eprint={2406.11903},
      archivePrefix={arXiv},
      primaryClass={q-fin.GN},
      url={https://arxiv.org/abs/2406.11903}, 
}

@ARTICLE{10818583,
  author={Yuan, Ziqiang and Wang, Kaiyuan and Zhu, Shoutai and Yuan, Ye and Zhou, Jingya and Zhu, Yanlin and Wei, Wenqi},
  journal={IEEE Transactions on Big Data}, 
  title={FinLLMs: A Framework for Financial Reasoning Dataset Generation With Large Language Models}, 
  year={2025},
  volume={11},
  number={5},
  pages={2264-2277},
  doi={10.1109/TBDATA.2024.3524083}}

@inproceedings{zhang2024multimodalfoundationagentfinancial,
      title={A multimodal foundation agent for financial trading: Tool-augmented, diversified, and generalist},
  author={Zhang, Wentao and Zhao, Lingxuan and Xia, Haochong and Sun, Shuo and Sun, Jiaze and Qin, Molei and Li, Xinyi and Zhao, Yuqing and Zhao, Yilei and Cai, Xinyu and others},
  booktitle={Proceedings of the 30th acm sigkdd conference on knowledge discovery and data mining},
  pages={4314--4325},
  year={2024}
}

@article{Wang2025,
   title={RETQA: A Large-Scale Open-Domain Tabular Question Answering Dataset for Real Estate Sector},
   volume={39},
   ISSN={2159-5399},
   url={http://dx.doi.org/10.1609/AAAI.V39I24.34734},
   DOI={10.1609/aaai.v39i24.34734},
   number={24},
   journal={Proceedings of the AAAI Conference on Artificial Intelligence},
   publisher={Association for the Advancement of Artificial Intelligence (AAAI)},
   author={Wang, Zhensheng and Yang, Wenmian and Zhou, Kun and Zhang, Yiquan and Jia, Weijia},
   year={2025},
   month=apr, pages={25452–25460} }

@misc{hussain2025artemisdaadvancedreasoningtransformation,
      title={ARTEMIS-DA: An Advanced Reasoning and Transformation Engine for Multi-Step Insight Synthesis in Data Analytics}, 
      author={Atin Sakkeer Hussain},
      year={2025},
      eprint={2412.14146},
      archivePrefix={arXiv},
      primaryClass={cs.AI},
      url={https://arxiv.org/abs/2412.14146}, 
}

@misc{zhu2025dianjinr1evaluatingenhancingfinancial,
      title={DianJin-R1: Evaluating and Enhancing Financial Reasoning in Large Language Models}, 
      author={Jie Zhu and Qian Chen and Huaixia Dou and Junhui Li and Lifan Guo and Feng Chen and Chi Zhang},
      year={2025},
      eprint={2504.15716},
      archivePrefix={arXiv},
      primaryClass={cs.AI},
      url={https://arxiv.org/abs/2504.15716}, 
}

@inproceedings{10.1145/3711896.3737157,
author = {Lu, Yihang and Xu, Yangyang and Qin, Qitao and Meng, Xianwei},
title = {TimeCapsule: Solving the Jigsaw Puzzle of Long-Term Time Series Forecasting with Compressed Predictive Representations},
year = {2025},
isbn = {9798400714542},
publisher = {Association for Computing Machinery},
address = {New York, NY, USA},
url = {https://doi.org/10.1145/3711896.3737157},
doi = {10.1145/3711896.3737157},
pages = {1987–1998},
numpages = {12},
location = {Toronto ON, Canada},
series = {KDD '25}
}

@misc{wang2023donotanswerdatasetevaluatingsafeguards,
      title={Do-Not-Answer: A Dataset for Evaluating Safeguards in LLMs}, 
      author={Yuxia Wang and Haonan Li and Xudong Han and Preslav Nakov and Timothy Baldwin},
      year={2023},
      eprint={2308.13387},
      archivePrefix={arXiv},
      primaryClass={cs.CL},
      url={https://arxiv.org/abs/2308.13387}, 
}

@inproceedings{es-etal-2024-ragas,
    title = "{RAGA}s: Automated Evaluation of Retrieval Augmented Generation",
    author = "Es, Shahul  and
      James, Jithin  and
      Espinosa Anke, Luis  and
      Schockaert, Steven",
    editor = "Aletras, Nikolaos  and
      De Clercq, Orphee",
    booktitle = "Proceedings of the 18th Conference of the European Chapter of the Association for Computational Linguistics: System Demonstrations",
    month = mar,
    year = "2024",
    address = "St. Julians, Malta",
    publisher = "Association for Computational Linguistics",
    url = "https://aclanthology.org/2024.eacl-demo.16/",
    doi = "10.18653/v1/2024.eacl-demo.16",
    pages = "150--158"
}

@inproceedings{min-etal-2023-factscore,
    title = "{FA}ct{S}core: Fine-grained Atomic Evaluation of Factual Precision in Long Form Text Generation",
    author = "Min, Sewon  and
      Krishna, Kalpesh  and
      Lyu, Xinxi  and
      Lewis, Mike  and
      Yih, Wen-tau  and
      Koh, Pang  and
      Iyyer, Mohit  and
      Zettlemoyer, Luke  and
      Hajishirzi, Hannaneh",
    editor = "Bouamor, Houda  and
      Pino, Juan  and
      Bali, Kalika",
    booktitle = "Proceedings of the 2023 Conference on Empirical Methods in Natural Language Processing",
    month = dec,
    year = "2023",
    address = "Singapore",
    publisher = "Association for Computational Linguistics",
    url = "https://aclanthology.org/2023.emnlp-main.741/",
    doi = "10.18653/v1/2023.emnlp-main.741",
    pages = "12076--12100"
}

@inproceedings{10.1145/3637528.3671629,
author = {Dong, Zihan and Fan, Xinyu and Peng, Zhiyuan},
title = {FNSPID: A Comprehensive Financial News Dataset in Time Series},
year = {2024},
isbn = {9798400704901},
publisher = {Association for Computing Machinery},
address = {New York, NY, USA},
url = {https://doi.org/10.1145/3637528.3671629},
doi = {10.1145/3637528.3671629},
booktitle = {Proceedings of the 30th ACM SIGKDD Conference on Knowledge Discovery and Data Mining},
pages = {4918–4927},
numpages = {10},
location = {Barcelona, Spain},
series = {KDD '24}
}

@inproceedings{pasupat-liang-2015-compositional,
    title = "Compositional Semantic Parsing on Semi-Structured Tables",
    author = "Pasupat, Panupong  and
      Liang, Percy",
    editor = "Zong, Chengqing  and
      Strube, Michael",
    booktitle = "Proceedings of the 53rd Annual Meeting of the Association for Computational Linguistics and the 7th International Joint Conference on Natural Language Processing (Volume 1: Long Papers)",
    month = jul,
    year = "2015",
    address = "Beijing, China",
    publisher = "Association for Computational Linguistics",
    url = "https://aclanthology.org/P15-1142/",
    doi = "10.3115/v1/P15-1142",
    pages = "1470--1480"
}

@misc{zhong2017seq2sqlgeneratingstructuredqueries,
      title={Seq2SQL: Generating Structured Queries from Natural Language using Reinforcement Learning}, 
      author={Victor Zhong and Caiming Xiong and Richard Socher},
      year={2017},
      eprint={1709.00103},
      archivePrefix={arXiv},
      primaryClass={cs.CL},
      url={https://arxiv.org/abs/1709.00103}, 
}

@inproceedings{kweon-etal-2023-open,
    title = "Open-{W}iki{T}able : Dataset for Open Domain Question Answering with Complex Reasoning over Table",
    author = "Kweon, Sunjun  and
      Kwon, Yeonsu  and
      Cho, Seonhee  and
      Jo, Yohan  and
      Choi, Edward",
    editor = "Rogers, Anna  and
      Boyd-Graber, Jordan  and
      Okazaki, Naoaki",
    booktitle = "Findings of the Association for Computational Linguistics: ACL 2023",
    month = jul,
    year = "2023",
    address = "Toronto, Canada",
    publisher = "Association for Computational Linguistics",
    url = "https://aclanthology.org/2023.findings-acl.526/",
    doi = "10.18653/v1/2023.findings-acl.526",
    pages = "8285--8297"
}

@inproceedings{chen-etal-2021-finqa,
    title = "{F}in{QA}: A Dataset of Numerical Reasoning over Financial Data",
    author = "Chen, Zhiyu  and
      Chen, Wenhu  and
      Smiley, Charese  and
      Shah, Sameena  and
      Borova, Iana  and
      Langdon, Dylan  and
      Moussa, Reema  and
      Beane, Matt  and
      Huang, Ting-Hao  and
      Routledge, Bryan  and
      Wang, William Yang",
    editor = "Moens, Marie-Francine  and
      Huang, Xuanjing  and
      Specia, Lucia  and
      Yih, Scott Wen-tau",
    booktitle = "Proceedings of the 2021 Conference on Empirical Methods in Natural Language Processing",
    month = nov,
    year = "2021",
    address = "Online and Punta Cana, Dominican Republic",
    publisher = "Association for Computational Linguistics",
    url = "https://aclanthology.org/2021.emnlp-main.300/",
    doi = "10.18653/v1/2021.emnlp-main.300",
    pages = "3697--3711"
}

@misc{kurisinkel2024text2timeseriesenhancingfinancialforecasting,
      title={Text2TimeSeries: Enhancing Financial Forecasting through Time Series Prediction Updates with Event-Driven Insights from Large Language Models}, 
      author={Litton Jose Kurisinkel and Pruthwik Mishra and Yue Zhang},
      year={2024},
      eprint={2407.03689},
      archivePrefix={arXiv},
      primaryClass={cs.CL},
      url={https://arxiv.org/abs/2407.03689}, 
}

@inproceedings{kong-etal-2025-time,
    title = "Time-{MQA}: Time Series Multi-Task Question Answering with Context Enhancement",
    author = "Kong, Yaxuan  and
      Yang, Yiyuan  and
      Hwang, Yoontae  and
      Du, Wenjie  and
      Zohren, Stefan  and
      Wang, Zhangyang  and
      Jin, Ming  and
      Wen, Qingsong",
    editor = "Che, Wanxiang  and
      Nabende, Joyce  and
      Shutova, Ekaterina  and
      Pilehvar, Mohammad Taher",
    booktitle = "Proceedings of the 63rd Annual Meeting of the Association for Computational Linguistics (Volume 1: Long Papers)",
    month = jul,
    year = "2025",
    address = "Vienna, Austria",
    publisher = "Association for Computational Linguistics",
    url = "https://aclanthology.org/2025.acl-long.1437/",
    doi = "10.18653/v1/2025.acl-long.1437",
    pages = "29736--29753",
    ISBN = "979-8-89176-251-0"
}

@misc{christakopoulou2024agentsthinkingfastslow,
      title={Agents Thinking Fast and Slow: A Talker-Reasoner Architecture}, 
      author={Konstantina Christakopoulou and Shibl Mourad and Maja Matarić},
      year={2024},
      eprint={2410.08328},
      archivePrefix={arXiv},
      primaryClass={cs.AI},
      url={https://arxiv.org/abs/2410.08328}, 
}

@inproceedings{qiao-etal-2024-autoact,
    title = "{A}uto{A}ct: Automatic Agent Learning from Scratch for {QA} via Self-Planning",
    author = "Qiao, Shuofei  and
      Zhang, Ningyu  and
      Fang, Runnan  and
      Luo, Yujie  and
      Zhou, Wangchunshu  and
      Jiang, Yuchen  and
      Lv, Chengfei  and
      Chen, Huajun",
    editor = "Ku, Lun-Wei  and
      Martins, Andre  and
      Srikumar, Vivek",
    booktitle = "Proceedings of the 62nd Annual Meeting of the Association for Computational Linguistics (Volume 1: Long Papers)",
    month = aug,
    year = "2024",
    address = "Bangkok, Thailand",
    publisher = "Association for Computational Linguistics",
    url = "https://aclanthology.org/2024.acl-long.165/",
    doi = "10.18653/v1/2024.acl-long.165",
    pages = "3003--3021"
}

@misc{li2025designingdomainspecificagentshierarchical,
      title={Designing Domain-Specific Agents via Hierarchical Task Abstraction Mechanism}, 
      author={Kaiyu Li and Jiayu Wang and Zhi Wang and Hui Qiao and Weizhan Zhang and Deyu Meng and Xiangyong Cao},
      year={2025},
      eprint={2511.17198},
      archivePrefix={arXiv},
      primaryClass={cs.AI},
      url={https://arxiv.org/abs/2511.17198}, 
}

@inproceedings{lin-etal-2025-llm,
    title = "On {LLM}-Based Scientific Inductive Reasoning Beyond Equations",
    author = "Lin, Brian S.  and
      Yuan, Jiaxin  and
      Zhou, Zihan  and
      Wang, Shouli  and
      Wang, Shuo  and
      Kong, Cunliang  and
      Shi, Qi  and
      Li, Yuxuan  and
      Yang, Liner  and
      Liu, Zhiyuan  and
      Sun, Maosong",
    editor = "Christodoulopoulos, Christos  and
      Chakraborty, Tanmoy  and
      Rose, Carolyn  and
      Peng, Violet",
    booktitle = "Proceedings of the 2025 Conference on Empirical Methods in Natural Language Processing",
    month = nov,
    year = "2025",
    address = "Suzhou, China",
    publisher = "Association for Computational Linguistics",
    url = "https://aclanthology.org/2025.emnlp-main.476/",
    doi = "10.18653/v1/2025.emnlp-main.476",
    pages = "9371--9394",
    ISBN = "979-8-89176-332-6"
}

@inproceedings{devlin2019bertpretrainingdeepbidirectional,
  title={{BERT}: Pre-training of Deep Bidirectional Transformers for Language Understanding},
  author={Devlin, Jacob and Chang, Ming-Wei and Lee, Kenton and Toutanova, Kristina},
  booktitle={Proceedings of the 2019 Conference of the North American Chapter of the Association for Computational Linguistics: Human Language Technologies, Volume 1 (Long and Short Papers)},
  pages={4171--4186},
  year={2019},
  publisher={Association for Computational Linguistics},
  doi={10.18653/v1/N19-1423},
  url={https://aclanthology.org/N19-1423/}
}

@misc{yang2025qwen3technicalreport,
      title={Qwen3 Technical Report}, 
      author={An Yang and Anfeng Li and Baosong Yang and Beichen Zhang and Binyuan Hui and Bo Zheng and Bowen Yu and Chang Gao and Chengen Huang and Chenxu Lv and Chujie Zheng and Dayiheng Liu and Fan Zhou and Fei Huang and Feng Hu and Hao Ge and Haoran Wei and Huan Lin and Jialong Tang and Jian Yang and Jianhong Tu and Jianwei Zhang and Jianxin Yang and Jiaxi Yang and Jing Zhou and Jingren Zhou and Junyang Lin and Kai Dang and Keqin Bao and Kexin Yang and Le Yu and Lianghao Deng and Mei Li and Mingfeng Xue and Mingze Li and Pei Zhang and Peng Wang and Qin Zhu and Rui Men and Ruize Gao and Shixuan Liu and Shuang Luo and Tianhao Li and Tianyi Tang and Wenbiao Yin and Xingzhang Ren and Xinyu Wang and Xinyu Zhang and Xuancheng Ren and Yang Fan and Yang Su and Yichang Zhang and Yinger Zhang and Yu Wan and Yuqiong Liu and Zekun Wang and Zeyu Cui and Zhenru Zhang and Zhipeng Zhou and Zihan Qiu},
      year={2025},
      eprint={2505.09388},
      archivePrefix={arXiv},
      primaryClass={cs.CL},
      url={https://arxiv.org/abs/2505.09388}, 
}

@misc{openai2025gptoss120bgptoss20bmodel,
      title={gpt-oss-120b \& gpt-oss-20b Model Card}, 
      author={Agarwal, Sandhini and Ahmad, Lama and Ai, Jason and Altman, Sam and Applebaum, Andy and Arbus, Edwin and Arora, Rahul K and Bai, Yu and Baker, Bowen and Bao, Haiming and others},
      year={2025},
      eprint={2508.10925},
      archivePrefix={arXiv},
      primaryClass={cs.CL},
      url={https://arxiv.org/abs/2508.10925}, 
}

@misc{wu2023timesnettemporal2dvariationmodeling,
      title={TimesNet: Temporal 2D-Variation Modeling for General Time Series Analysis}, 
      author={Haixu Wu and Tengge Hu and Yong Liu and Hang Zhou and Jianmin Wang and Mingsheng Long},
      year={2023},
      eprint={2210.02186},
      archivePrefix={arXiv},
      primaryClass={cs.LG},
      url={https://arxiv.org/abs/2210.02186}, 
}

@inproceedings{NEURIPS2024_0113ef46,
 author = {Wang, Yuxuan and Wu, Haixu and Dong, Jiaxiang and Qin, Guo and Zhang, Haoran and Liu, Yong and Qiu, Yunzhong and Wang, Jianmin and Long, Mingsheng},
 booktitle = {Advances in Neural Information Processing Systems},
 doi = {10.52202/079017-0015},
 editor = {A. Globerson and L. Mackey and D. Belgrave and A. Fan and U. Paquet and J. Tomczak and C. Zhang},
 pages = {469--498},
 publisher = {Curran Associates, Inc.},
 title = {TimeXer: Empowering Transformers for Time Series Forecasting with Exogenous Variables},
 url = {https://proceedings.neurips.cc/paper_files/paper/2024/file/0113ef4642264adc2e6924a3cbbdf532-Paper-Conference.pdf},
 volume = {37},
 year = {2024}
}

@inproceedings{ICLR2024_a7ac8a21,
 author = {Wang, Shiyu and Wu, Haixu and Shi, Xiaoming and Hu, Tengge and Luo, Huakun and Ma, Lintao and Zhang, James and ZHOU, JUN},
 booktitle = {International Conference on Learning Representations},
 editor = {B. Kim and Y. Yue and S. Chaudhuri and K. Fragkiadaki and M. Khan and Y. Sun},
 pages = {38626--38652},
 title = {TimeMixer: Decomposable Multiscale Mixing for Time Series Forecasting},
 url = {https://proceedings.iclr.cc/paper_files/paper/2024/file/a7ac8a21e5a27e7ab31a5f42a0117bdb-Paper-Conference.pdf},
 volume = {2024},
 year = {2024}
}

@inproceedings{liu2024itransformerinvertedtransformerseffective,
      title={itransformer: Inverted transformers are effective for time series forecasting},
  author={Liu, Yong and Hu, Tengge and Zhang, Haoran and Wu, Haixu and Wang, Shiyu and Ma, Lintao and Long, Mingsheng},
  booktitle={International conference on learning representations},
  volume={2024},
  pages={11116--11140},
  year={2024}
}

@inproceedings{shi2025timemoebillionscaletimeseries,
      title={Time-moe: Billion-scale time series foundation models with mixture of experts},
  author={Shi, Xiaoming and Wang, Shiyu and Nie, Yuqi and Li, Dianqi and Ye, Zhou and Wen, Qingsong and Jin, Ming},
  booktitle={International conference on learning representations},
  volume={2025},
  pages={34635--34667},
  year={2025}
}

@misc{zeng2025futurexadvancedlivebenchmark,
      title={FutureX: An Advanced Live Benchmark for LLM Agents in Future Prediction}, 
      author={Zhiyuan Zeng and Jiashuo Liu and Siyuan Chen and Tianci He and Yali Liao and Yixiao Tian and Jinpeng Wang and Zaiyuan Wang and Yang Yang and Lingyue Yin and Mingren Yin and Zhenwei Zhu and Tianle Cai and Zehui Chen and Jiecao Chen and Yantao Du and Xiang Gao and Jiacheng Guo and Liang Hu and Jianpeng Jiao and Xiangsheng Li and Jingkai Liu and Shuang Ni and Zhoufutu Wen and Ge Zhang and Kaiyuan Zhang and Xin Zhou and Jose Blanchet and Xipeng Qiu and Mengdi Wang and Wenhao Huang},
      year={2025},
      eprint={2508.11987},
      archivePrefix={arXiv},
      primaryClass={cs.AI},
      url={https://arxiv.org/abs/2508.11987}, 
}

@inproceedings{karger2025forecastbenchdynamicbenchmarkai,
      title={Forecastbench: A dynamic benchmark of ai forecasting capabilities},
  author={Karger, Ezra and Bastani, Houtan and Yueh-Han, Chen and Jacobs, Zachary and Halawi, Danny and Zhang, Fred and Tetlock, Philip E},
  booktitle={International Conference on Learning Representations},
  volume={2025},
  pages={93943--93980},
  year={2025}
}

@misc{yuan2025forecastfutureoutcomereasoning,
      title={FOReCAst: The Future Outcome Reasoning and Confidence Assessment Benchmark}, 
      author={Zhangdie Yuan and Zifeng Ding and Andreas Vlachos},
      year={2025},
      eprint={2502.19676},
      archivePrefix={arXiv},
      primaryClass={cs.LG},
      url={https://arxiv.org/abs/2502.19676}, 
}

@inproceedings{Chen_2024,
   title={FinTextQA: A Dataset for Long-form Financial Question Answering},
   url={http://dx.doi.org/10.18653/v1/2024.acl-long.328},
   DOI={10.18653/v1/2024.acl-long.328},
   booktitle={Proceedings of the 62nd Annual Meeting of the Association for Computational Linguistics (Volume 1: Long Papers)},
   publisher={Association for Computational Linguistics},
   author={Chen, Jian and Zhou, Peilin and Hua, Yining and Xin, Loh and Chen, Kehui and Li, Ziyuan and Zhu, Bing and Liang, Junwei},
   year={2024},
   pages={6025–6047} }

@misc{qiu2024tqabenchevaluatingllmsmultitable,
      title={TQA-Bench: Evaluating LLMs for Multi-Table Question Answering with Scalable Context and Symbolic Extension}, 
      author={Zipeng Qiu and You Peng and Guangxin He and Binhang Yuan and Chen Wang},
      year={2024},
      eprint={2411.19504},
      archivePrefix={arXiv},
      primaryClass={cs.AI},
      url={https://arxiv.org/abs/2411.19504}, 
}

@inproceedings{10.1145/3768292.3770361,
author = {Choi, Chanyeol and Kwon, Jihoon and Ha, Jaeseon and Choi, Hojun and Kim, Chaewoon and Lee, Yongjae and Sohn, Jy-yong and Lopez-Lira, Alejandro},
title = {FinDER: Financial Dataset for Question Answering and Evaluating Retrieval-Augmented Generation},
year = {2025},
isbn = {9798400722202},
publisher = {Association for Computing Machinery},
address = {New York, NY, USA},
url = {https://doi.org/10.1145/3768292.3770361},
doi = {10.1145/3768292.3770361},
booktitle = {Proceedings of the 6th ACM International Conference on AI in Finance},
pages = {638–646},
numpages = {9},
location = {
},
series = {ICAIF '25}
}

@misc{google2026gemma4modelcard,
  title        = {Gemma 4 Model Card},
  author       = {{Google DeepMind}},
  year         = {2026},
  howpublished = {\url{https://ai.google.dev/gemma/docs/core/model_card_4}},
  note         = {Accessed: 2026-05-14}
}

\appendix

\section{Experimental Results Details}
\subsection{Implementation Details}
The full system is implemented in PyTorch 2.5.1\footnote{\url{https://pytorch.org/}}. The agent modules are developed on top of the LangGraph (0.2.73) and LangChain (0.3.17) frameworks to support multi-step LLM workflows with controllable agent execution logic. The SLU module is obtained by fine-tuning a pretrained BERT model\footnote{\url{https://huggingface.co/google-bert/bert-base-cased}}. During training, we optimize model parameters using the Adam optimizer for 10 epochs. The overall objective consists of two components: intent forecast is trained with binary cross-entropy (BCE) loss, while slot tagging is trained with cross-entropy loss. After training, we run SLU tag forecasting on a single GeForce RTX 4090 GPU to provide structured semantic signals for downstream modules.

For the time-series forecast models, we optimize the parameters using the Adam optimizer and set the maximum training budget to 50 epochs. To mitigate overfitting and avoid unnecessary updates once validation performance saturates, we adopt an early-stopping strategy: training is terminated when the validation loss fails to improve for three consecutive epochs (patience:3).

Moreover, we construct training instances via a sliding-window scheme, where each example maps 49 historical observations to 1 future observation, yielding “49 to 1” single-step forecast samples (650 windows in total under the current configuration). During inference and evaluation, we further perform multi-step forecasting in an autoregressive manner by setting the forecast horizon to 15 steps (horizon: 15) and iteratively rolling the input window forward to generate forecasts for the subsequent 15 time steps.

To evaluate the performance of all large language models (LLMs), we employ a cluster of fourteen NVIDIA A800-SXM4-80GB GPUs. Specifically, two GPUs are allocated to Qwen3 30B, two GPUs are allocated to GEMMA 4-31B, eight GPUs are allocated to GPT-oss-120B, and the remaining two GPUs are allocated to GPT-oss-20B.

\subsection{Experimental Results of SQL Accuracy}
\label{sec:appendixA.2}

This section reports SQL query accuracy for three task types: query reasoning (QR), numerical forecasting (NF), and forecast-based reasoning (FR). Since QR queries require both database access and more complex reasoning, the corresponding SQL statements are typically more sophisticated; in contrast, NF and FR queries only need to retrieve historical data.

Experimental results show that, for SQL generation, History2SQL (H2S) achieves a substantially higher average pass@1 (99.40\%; see Table~\ref{Table 6}) than Query2SQL (Q2S) (71.23\%). In particular, the correctness of query reasoning (QR) is determined directly by SQL execution outcomes, whereas the accuracy of numerical forecasting (NF) and forecast-based reasoning (FR) largely depends on the retrieval quality provided by H2S. Concretely, higher-quality retrieval generally leads to lower MRE on NF and higher accuracy on FR.

\begin{table}
  \centering
  \small
  \begin{tabular}{lcc}
    \hline
    & \multicolumn{2}{c}{\textbf{History SQL}} \\
    \cmidrule(lr){2-3}
    \textbf{Model} & \textbf{ECR} & \textbf{Pass@1} \\
    \hline
    Qwen3 30B      & 100\% & 99.05\% \\
    \hline
    GPT-oss-20B    & 100\% & 99.25\% \\
    \hline
    GPT-oss-120B   & 100\% & 99.60\% \\
    \hline
    \textbf{GEMMA 4-31B}       & \textbf{100\%} & \textbf{99.70\%} \\
    \hline
  \end{tabular}
  \caption{\label{Table 6}
    History SQL ECR and Pass@1.}
\end{table}

\subsection{Detailed Results of the SLU Module}
\label{sec:appendixA.3}

\begin{table*}[t]
  \centering
  \small
  \setlength{\tabcolsep}{4pt}
  \begin{tabular}{llccccccc cccc}
    \hline
        \multirow{2}{*}{\textbf{Model}} &
        \multirow{2}{*}{\textbf{Setting}} &
        \multicolumn{4}{c}{\textbf{QR}} &
        \multicolumn{3}{c}{\textbf{NF}} &
        \multicolumn{4}{c}{\textbf{FR}} \\
        \cmidrule(lr){3-6} \cmidrule(lr){7-9} \cmidrule(lr){10-13}
        & & \textbf{P} & \textbf{R} & \textbf{F1} & \textbf{Acc}
          & \textbf{MSE} & \textbf{MAE} & \textbf{MRE}
          & \textbf{P} & \textbf{R} & \textbf{F1} & \textbf{Acc} \\
        \hline

    \multirow{2}{*}{Qwen3 30B}
      & SQFRS   & 0.7006 & 0.7003 & 0.7000 & 0.6924 & 81.3165 & 2.4583 & 0.0617 & 0.5539 & 0.5532 & 0.5536 & 0.5529 \\
      & W/O SLU & 0.6571 & 0.6523 & 0.6529 & 0.6486 & 116.9249 & 2.7243 & 0.0704 & 0.5123 & 0.5166 & 0.5145 & 0.5029 \\
    \hline
    \multirow{2}{*}{GPT-oss-20B}
      & SQFRS   & 0.7236 & 0.7188 & 0.7186 & 0.7152 & 81.3120 & 2.4561 & 0.0616 & 0.5959 & 0.6535 & 0.6234 & 0.5929 \\
      & W/O SLU & 0.6771 & 0.6730 & 0.6734 & 0.6705 & 116.8996 & 2.7237 & 0.0704 & 0.5531 & 0.5800 & 0.5762 & 0.5509 \\
    \hline

    \multirow{2}{*}{GPT-oss-120B}
      & SQFRS   & 0.7314 & 0.7316 & 0.7309 & 0.7178
               & \textbf{80.5744} & \textbf{2.4140} & \textbf{0.0590}
               & 0.6160 & 0.6596 & 0.6319 & 0.6171 \\
      & W/O SLU & 0.6800 & 0.6794 & 0.6773 & 0.6738
               & 114.5288 & 2.7062 & 0.0700
               & 0.5769 & 0.6020 & 0.5884 & 0.5671 \\
    \hline

    \multirow{2}{*}{GEMMA 4-31B}
      & \textbf{SQFRS}   & \textbf{0.7372} & \textbf{0.7369} & \textbf{0.7370} & \textbf{0.7238} & 81.4815 & 2.4828 & 0.0623 & \textbf{0.6752} & \textbf{0.6844} & \textbf{0.6798} & \textbf{0.6243} \\
      & W/O SLU & 0.6707 & 0.6678 & 0.6690 & 0.6532 & 117.2105 & 2.7399 & 0.0708 & 0.5958 & 0.6587 & 0.5759 & 0.5700 \\
    \hline
  \end{tabular}

  \caption{\label{Table 7} Experimental Results of the SLU Module.}
\end{table*}

This section further substantiates the contribution of the SLU module with more fine-grained experimental evidence. First, we compare SQFRS with the baseline method of disabling the SLU module (W/O SLU). The results are shown in Table \ref {Table 7}. Compared to the W/O SLU baseline in terms of averaged metrics (QR: F1 66.82\%, Acc 66.15\%; NF: 0.0704; FR: Acc 56.38\%, F1 54.77\%), SQFRS delivers consistent improvements across all three tasks (QR: F1 72.16\%, Acc 71.23\%; NF: 0.0612; FR: F1 62.22\%, Acc 59.68\%). Moreover, due to space constraints in the main paper, we report only a subset of metrics (Acc and MRE) in the main text. Therefore, we have added more results (F1, MSE and MAE) in the appendix to support more comprehensive experimental analysis. Taken together, the results suggest that intent labels facilitate more appropriate workflow allocation in the DRA module, while slot labels impose more precise constraints on SQL generation. In combination, these signals highlight the critical role of the SLU module in improving end-to-end performance, consistent with the conclusions in the main text.

\subsection{Robustness Analysis via In‑Context Learning}
\label{sec:appendixA.4}

\begin{table}[t]
  \centering
  \small
  \begin{tabular}{lccccc}
    \hline
    \multicolumn{1}{c}{\multirow{2}{*}{\textbf{Model}}} & 
    \multirow{2}{*}{\textbf{Method}} & 
    \multicolumn{1}{c}{\textbf{QR}} &
    \multicolumn{1}{c}{\textbf{NF}} & 
    \multicolumn{1}{c}{\textbf{FR}} \\
    \cmidrule(lr){3-3} \cmidrule(lr){4-5}
    & & \textbf{Acc} & \textbf{MRE} & \textbf{Acc} \\
    \hline
    \multirow{2}{*}{Qwen3 30B} 
      & SQFRS   & 0.6924 & 0.0617 & 0.5529 \\
      & LLM-SLU & 0.6810 & 0.0685 & 0.5400 \\
    \hline
    \multirow{2}{*}{GPT-oss-20B} 
      & SQFRS   & 0.7152 & 0.0616 & 0.5929 \\
      & LLM-SLU & 0.7029 & 0.0678 & 0.5886 \\
    \hline
    \multirow{2}{*}{GPT-oss-120B} 
      & SQFRS   & 0.7178 & 0.0590 & 0.6171\\
      & LLM-SLU & 0.7120 & 0.0675 & 0.6029 \\
    \hline
    \multirow{2}{*}{GEMMA 4-31B}  
      & SQFRS   & 0.7238 & 0.0623 & 0.6243 \\
      & LLM-SLU & 0.7073 & 0.0706 & 0.6071 \\
    \hline
  \end{tabular}
  \caption{\label{Table 8}Robustness analysis via in‑context learning.}
\end{table}

To further verify the robustness of the SLU module, we conduct an additional experiment under an in-context learning (ICL) setting. In this configuration, instead of using a separately fine-tuned BERT model, we directly leverage the LLM itself to generate SLU labels via few-shot prompting, thereby avoiding any extra training; the detailed results are reported in Table \ref{Table 8}. The experimental results show that, for NF-type tasks, the mean relative error (MRE) of LLM-SLU is generally higher than that of the proposed SQFRS method. For QR and FR tasks, LLM-SLU also performs slightly worse than SQFRS. This indicates that although the LLM can generate SLU labels via few-shot prompting without additional training, its numerical information extraction capability remains less stable when interfacing with external time-series forecasting models. Meanwhile, its ability to perform further reasoning based on predicted values is still limited. As a result, LLM-SLU achieves lower final accuracy (ACC) on the overall tasks than the BERT-fine-tuned SQFRS method.

Furthermore, the comparative results show that the ICL-based SLU approach (LLM-SLU) achieves an average QR accuracy of 70.08\%, an NF MRE of 0.0686, and an FR accuracy of 58.47\%. Compared with the BERT-fine-tuned SQFRS model, LLM-SLU performs slightly worse on NF-type tasks, while its accuracy on QR and FR tasks is consistently lower than that of SQFRS. Overall, the experimental results across all backbone models consistently confirm the stability and reliability of this conclusion.

\subsection{Fine-Grained Analysis of Forecast-Based Question Answering}

To further evaluate models' reasoning capabilities over forecasted data, we categorize the test instances into three types according to the required reasoning operations: \textit{threshold comparison}, \textit{extreme-value reasoning}, and \textit{multi-period trend comparison}. Threshold comparison primarily involves direct numerical comparison; extreme-value reasoning additionally requires temporal alignment and the extraction of target values from forecast sequences; and multi-period trend comparison further requires models to align and jointly reason over values across multiple future periods. Detailed experimental results are presented in Table \ref{Table 9}.

\begin{table}[t]
  \centering
  \small
  \setlength{\tabcolsep}{4pt}
  \begin{tabular}{lccc}
    \hline
    \textbf{Model} &
    \textbf{Threshold} &
    \begin{tabular}[c]{@{}c@{}}\textbf{Extreme}\\\textbf{Value}\end{tabular} &
    \begin{tabular}[c]{@{}c@{}}\textbf{Trend}\\\textbf{Comparison}\end{tabular} \\
    \hline
    Qwen3-30B    & 76.0\% & 44.5\% & 54.8\% \\
    GPT-oss-20B  & 66.0\% & 56.0\% & 55.2\% \\
    GPT-oss-120B & 80.0\% & 76.0\% & 55.6\% \\
    GEMMA 4-31B  & 87.0\% & 82.5\% & 52.8\% \\
    \hline
  \end{tabular}
  \caption{Fine-grained results on different types of forecast-based reasoning questions.}
  \label{Table 9}
\end{table}

The results show that model performance decreases significantly as reasoning complexity increases. On threshold comparison tasks, the four models achieve accuracies ranging from 66\% to 87\%, indicating a relatively strong capability for direct numerical comparison. When temporal alignment and value extraction are additionally required in extreme-value reasoning, the accuracies of Qwen3-30B and GPT-oss-20B decrease by 31.5 and 10.0 percentage points, respectively. In contrast, GPT-oss-120B and GEMMA 4-31B exhibit smaller declines of 4.0 and 4.5 percentage points, respectively. On the more challenging multi-period trend comparison tasks, all models achieve accuracies of only 52.8\%--55.6\%. These results suggest that current models can handle direct numerical comparisons relatively well but still face considerable challenges in compositional reasoning tasks that jointly involve temporal alignment, value extraction, and cross-period comparison.

\section{Selection of Time-Series Forecasting Models}
\label{sec:appendixB}

This section presents two sets of comparative experiments—on the original dataset and on the STQA dataset constructed in this work—to evaluate the effectiveness of alternative time-series forecasting models and thereby identify the optimal choice. To ensure a controlled and fair comparison, we adopt a fixed workflow setting and conduct ablation studies separately for the NF and FR tasks. In total, we consider six forecasting approaches, including a general-purpose large language model (GPT-oss-120B), a state-of-the-art LLM-based model (Time-MoE), and four lightweight time-series models (TimesNet, I-Transformer, TimeMixer, and TimeXer).

Recent studies suggest that forecasting trading volume trends requires not only modeling recent trading sequences, but also jointly capturing heterogeneous market information such as inter-stock relations, long-term trends, and unexpected events. Since short-term trading data are often affected by random fluctuations and sudden information shocks, their predictive accuracy can be substantially compromised. Motivated by this, we do not directly treat trading volume as a stable forecasting target. Instead, we select four market variables with stronger structural characteristics—open price, close price, high price, and low price—and construct both numeric forecasting tasks and forecast-based reasoning QA tasks.

On the original dataset, we focus on the NF and FR tasks and benchmark the selected time-series models against the best-performing LLM baseline within our evaluation scope, with results reported in Table \ref{Table 10}. The results show that, among the time-series baselines, Time-MoE achieves the best average performance (open: MRE = 0.0686, close: MRE = 0.0799, high: MRE = 0.0710, low: MRE = 0.0717), whereas I-Transformer incurs the largest forecast errors (open: MRE = 0.1154, close: MRE = 0.0998, high: MRE = 0.0695, low: MRE = 0.1167). These findings indicate that Time-MoE provides stronger numerical forecasting capability and are consistent with the model-selection conclusions in the main text.

From the NF (numeric forecasting) perspective, Time-MoE consistently yields the lowest error across all LLM backbone configurations considered in this work (e.g., MRE = 0.0590 when paired with GPT-oss-120B), whereas I-Transformer exhibits substantially higher errors (MRE = 0.1035). Under the more challenging FR (forecast-based reasoning) setting, which requires forecast-driven reasoning, Time-MoE likewise achieves higher accuracy, and this advantage remains stable across backbone configurations (e.g., Acc = 62.43\% when paired with GEMMA 4-31B), with results reported in Table \ref{Table 11}.

In addition, throughout this paper, the best result attained by each method is highlighted in bold in all experimental tables.

Overall, across both the original dataset and the STQA dataset, Time-MoE consistently emerges as the best-performing time-series forecast model. Its performance consistency across datasets and backbone configurations further supports the stability and reliability of this conclusion.

\begin{table*}
  \centering
  \small
  \begin{tabular}{cccccccc}
    \hline
    \textbf{Price Fields} & \textbf{Forecast-15days} & \textbf{TimeXer} & \textbf{I-Transformer} & \textbf{TimeMixer} & \textbf{TimesNet} & \textbf{GPT-oss-120B} & \textbf{Time-MoE} \\ 
    \hline
    \multirow{3}{*}{open} 
    & MSE & 252.2331 & 474.3625 & 147.4776 & 202.2346 & 202.2304 & \textbf{145.3490} \\
    & MAE & 4.1762 & 5.5079 & 3.4352 & 3.8685 & 4.0777 & \textbf{3.3174} \\
    & MRE & 0.0868 & 0.1154 & 0.0727 & 0.0810 & 0.0876 & \textbf{0.0686} \\
    \hline

    \multirow{3}{*}{close} 
    & MSE & 216.6427 & 289.5159 & 364.7470 & 673.4907 & 258.6725 & \textbf{202.9997} \\
    & MAE & 4.0625 & 4.7266 & 5.0233 & 7.0094 & 4.4143 & \textbf{3.7858} \\
    & MRE & 0.0863 & 0.0998 & 0.1050 & 0.1464 & 0.0929 & \textbf{0.0799} \\
    \hline

    \multirow{3}{*}{high} 
    & MSE & 296.3000 & 161.8028 & \textbf{129.8486} & 214.8473 & 234.4453 & 168.7912 \\
    & MAE & 4.5869 & 3.2970 & \textbf{3.2490} & 4.0043 & 4.2024 & 3.4215 \\
    & MRE & 0.0958 & 0.0695 & \textbf{0.0688} & 0.0833 & 0.0892 & 0.0710 \\
    \hline

    \multirow{3}{*}{low} 
    & MSE & 253.6796 & 479.4563 & 178.1207 & 203.3644 & 262.8046 & \textbf{168.3481} \\
    & MAE & 4.1818 & 5.5366 & 3.7305 & 3.8966 & 4.4811 & \textbf{3.4269} \\
    & MRE & 0.0876 & 0.1167 & 0.0792 & 0.0820 & 0.0943 & \textbf{0.0717} \\
    \hline
  \end{tabular}
  \caption{\label{Table 10}
    Experimental Results of the Time Series Forecasting Model Based on Raw Data.}
\end{table*}

\begin{table*}
  \centering
  \small
  \begin{tabular}{llccccccc}
    \hline
    \multicolumn{1}{c}{\multirow{2}{*}{\textbf{Model}}} & 
    {\multirow{2}{*}{\textbf{Method}}}
      & \multicolumn{4}{c}{\textbf{FR}}
      & \multicolumn{3}{c}{\textbf{NF}} \\
    \cline{3-6}\cline{7-9}
      &  & \textbf{P} & \textbf{R} & \textbf{F1} & \textbf{Acc}
         & \textbf{MSE} & \textbf{MAE} & \textbf{MRE} \\
    \hline
    \multirow{5}{*}{Qwen3 30B} & TimeXer     & 0.4768 & 0.4839 & 0.4803 & 0.4743 & 178.0094 & 2.8938 & 0.0855 \\
                               & I-Transformer     & 0.5536 & 0.5529 & 0.5515 & 0.5523 & 425.2180 & 4.3682 & 0.1049 \\
                               & TimeMixer     & 0.4847 & 0.4901 & 0.4874 & 0.4843 & 161.6847 & 2.8681 & 0.0819 \\
                               & TimesNet    & 0.5288 & 0.5334 & 0.5311 & 0.5219 & 114.4559 & 2.6444 & 0.0698 \\
                               & \textbf{TimesMoE}    & \textbf{0.5539} & \textbf{0.5532} & \textbf{0.5536} & \textbf{0.5529} & \textbf{81.3165} & \textbf{2.4583} & \textbf{0.0617} \\
    \hline
    \multirow{5}{*}{GPT-oss-20B} & TimeXer     & 0.5409 & 0.5504 & 0.5435 & 0.5314 & 203.9106 & 3.0325 & 0.0883 \\
                               & I-Transformer     & 0.5796 & 0.6096 & 0.6046 & 0.5836 & 424.5789 & 4.3618 & 0.1047 \\
                               & TimeMixer     & 0.5661 & 0.5960 & 0.5821 & 0.5557 & 179.4454 & 2.9255 & 0.0857 \\
                               & TimesNet    & 0.5746 & 0.5976 & 0.5924 & 0.5700 & 115.0367 & 2.7085 & 0.0701 \\
                               & \textbf{TimesMoE}    & \textbf{0.5959} & \textbf{0.6535} & \textbf{0.6234} & \textbf{0.5929} & \textbf{81.3120} & \textbf{2.4561} & \textbf{0.0616} \\
    \hline
    \multirow{5}{*}{GPT-oss-120B} & TimeXer     & 0.5847 & 0.5982 & 0.5929 & 0.5714 & 164.4769 & 2.8688 & 0.0822 \\
                               & I-Transformer     & 0.6009 & 0.6412 & 0.6248 & 0.6000 & 422.3081 & 4.2847 & 0.1035 \\
                               & TimeMixer     & 0.5849 & 0.6023 & 0.5986 & 0.5729 & 146.8535 & 2.8449 & 0.0797 \\
                               & TimesNet    & 0.5926 & 0.6371 & 0.6143 & 0.5914 & 114.9482 & 2.6492 & 0.0699 \\
                               & \textbf{TimesMoE}    & \textbf{0.6160} & \textbf{0.6596} & \textbf{0.6319} & \textbf{0.6171} & \textbf{80.5744} & \textbf{2.4140} & \textbf{0.0590} \\
    \hline
    \multirow{5}{*}{GEMMA 4-31B} & TimeXer     & 0.6542 & 0.6774 & 0.6558 & 0.6187 & 182.5206 & 2.9270 & 0.0904 \\
                               & I-Transformer     & 0.6736 & 0.6839 & 0.6787 & 0.6237 & 487.8556 & 4.6166 & 0.1135 \\
                               & TimeMixer     & 0.6703 & 0.6819 & 0.6761 & 0.6206 & 167.6567 & 2.8761 & 0.0823 \\
                               & TimesNet    & 0.6717 & 0.6828 & 0.6767 & 0.6214 & 122.7855 & 2.7601 & 0.0725 \\
                               & \textbf{TimesMoE}    & \textbf{0.6752} & \textbf{0.6844} & \textbf{0.6798} & \textbf{0.6243} & \textbf{81.4815} & \textbf{2.4828} & \textbf{0.0623} \\
    \hline
  \end{tabular}
  \caption{\label{Table 11}
    Comparative Experiment of Time Series Forecasting Models and Large Language Model.}
\end{table*}

\section{Statistics}
\label{sec:appendixC}

Appendix C reports the categorization and descriptive statistics of the QA data in the Dataset Metrics Statistics Table, with details summarized as follows (see Table \ref{Table 12}). The STQA dataset spans three domains—query reasoning, numeric forecasting, and forecast-based reasoning—and is instantiated from 72 templates, including 42 templates for query-oriented questions and 30 templates for forecasting-oriented questions; the latter further comprises 16 templates for numeric forecasting and 14 templates for forecast-based reasoning. By filling slots defined by these 72 template types, we construct STQA with 4,417 tables and 31,400 QA pairs, and split the data by tables into a training set of 3,094 tables (21,980 QA pairs), a validation set of 879 tables (6,280 QA pairs), and a test set of 444 tables (3,140 QA pairs).  We additionally define 10 intent categories and 5 slot categories: the intent set includes three query-reasoning intents, three numeric-forecasting intents, and four forecast-based reasoning intents; in total, 26 question types are covered, consisting of 7,250 reasoning QA pairs and 8,000 comparison QA pairs, with the remaining 15,250 QA pairs categorized as direct-type questions (i.e., direct query or direct forecast questions). 

Following the template-based domain partition, STQA contains 10,500 query-reasoning QA pairs, 10,400 numeric-forecasting QA pairs, and 10,500 forecast-based reasoning QA pairs. Within the query-reasoning subset, we analyze the data from three complementary perspectives. First, by table scope, 8,000 QA pairs are single-table instances, whereas 2,500 QA pairs involve cross-table reasoning. Second, by temporal scope, 8,000 QA pairs are associated with a single timestamp, while 2,500 QA pairs span multiple timestamps. In addition, 3,250 QA pairs are annotated as ambiguous-time queries, where the temporal constraint is underspecified (e.g., relative expressions such as “the past three days” or “a week ago”). The query-reasoning subset can also be decomposed by question nature. Under this view, 6,750 QA pairs correspond to numeric-query questions, while 3,750 QA pairs correspond to query-reasoning questions. For the numeric-forecast subset, all 10,400 QA pairs are single-table and single-timestamp instances, among which 6,000 QA pairs involve ambiguous-time forecasting. Likewise, for the forecast-based reasoning subset, all 10,500 QA pairs are single-table, multi-timestamp instances designed to evaluate models’ ability to reason based on forecasted results.

\begin{table*}
  \centering
  \small
  \begin{tabular}{lc}
    \hline
    \textbf{Label} & \textbf{Statistics num} \\
    \hline
    Number of templates & 72  \\
    Query template count & 42  \\
    Forecast template count & 30 \\
    Number of QA pairs & 31400 \\
    Single stock QA quantity & 28000  \\
    Forecast QA quantity & 20900 \\
    Number of multi-stock QAs & 2500  \\
    Number of time period QAs & 19000 \\
    Number of fuzzy time QAs & 18500 \\
    Domain & 3 \\
    Intent & 10 \\
    Forecast intent & 7 \\
    Query intent & 3 \\
    Slot & 5 \\
    Numerical Query QA & 6750 \\
    Query Reasoning QA & 3750 \\
    Numerical Forecast QA & 10400 \\
    forecast-based Reasoning QA & 10500 \\
    Question Types & 26 \\
    Number of Stocks & 4417 \\
    Test set QA & 3140 \\
    Number of stocks in the test set & 444 \\
    Validation set QA & 6280 \\
    Number of stocks in the validation set & 879 \\
    Training set QA & 21980 \\
    Number of stocks in the training set & 3094 \\
    Comparative problem QA & 10500 \\
    Query QA count & 10500 \\
    Query class single-stock QA count & 8000 \\
    Query class multi-stock QA count & 2500 \\
    Query class single-time QA count & 8000  \\
    Query class multi-time QA count & 2500  \\
    Query class fuzzy-time QA count & 3250 \\
    Query class time-range QA count & 2500 \\
    Forecast class single-table forecast QA count & 10400 \\
    forecast-based reasoning single-table forecast QA count & 10500 \\
    Forecast-type fuzzy date forecast QA count & 7150 \\
    forecast-based reasoning fuzzy date forecast QA count & 7500 \\
    forecast-based reasoning time period QA count & 9000 \\
    \hline
  \end{tabular}
  \caption{\label{Table 12}
    Dataset Metrics Statistics Table.}
\end{table*}

\section{Bottleneck Exploration}
\label{sec:appendixD}
Appendix D reports the experimental results of the SQFRS framework under multiple Ground-Truth (GT) settings, covering Ground-Truth SLU (Table \ref{Table 13}), Ground-Truth Workflow (Table \ref{Table 14}), Ground-Truth History SQL (Table \ref{Table 15}), and Ground-Truth FD (Table \ref{Table 16}). Overall, incorporating GT supervision signals yields consistent performance improvements; however, the magnitude of these gains differs substantially across modules.

\begin{table*}
  \centering
  \small
  \begin{tabular}{lccccccccccc}
    \hline
    & \multicolumn{4}{c}{\textbf{QR}} &\multicolumn{3}{c}{\textbf{NF}} & \multicolumn{4}{c}{\textbf{FR}}   \\
    \cmidrule(lr){2-5} \cmidrule(lr){6-8} \cmidrule(lr){9-12}
    \textbf{Model} & \textbf{P} & \textbf{R} & \textbf{F1} & \textbf{Acc} & \textbf{MSE} & \textbf{MAE} & \textbf{MRE} & \textbf{P} & \textbf{R} & \textbf{F1} & \textbf{Acc} \\
    \hline
    Qwen3 30B & 0.7105 & 0.7091 & 0.7085 & 0.6990 & 81.312530 & 2.4576 & 0.061611 & 0.5676 & 0.5662 & 0.5683 & 0.5586 \\
    \hline
    GPT-oss-20B & 0.7317 & 0.7300 & 0.7298 & 0.7286 & 81.226003 & 2.4528 & 0.061578 & 0.6255 & 0.6579 & 0.6414 & 0.6100 \\
    \hline
    GPT-oss-120B & 0.7407 & 0.7409 & 0.7402 & 0.7381 & 80.348970 & 2.4127 & 0.058952 & 0.6472 & 0.6619 & 0.6607 & 0.6234 \\
    \hline
    GEMMA 4-31B & 0.7412 & 0.7442 & 0.7409 & 0.7392 & 81.404654 & 2.4823 & 0.062075 & 0.6821 & 0.6916 & 0.6816 & 0.6357 \\
    \hline
  \end{tabular}
  \caption{\label{Table 13}
    Ground Truth SLU.}
\end{table*}

\begin{table*}
  \centering
  \small
  \begin{tabular}{lccccccccccc}
    \hline
    & \multicolumn{4}{c}{\textbf{QR}} &\multicolumn{3}{c}{\textbf{NF}} & \multicolumn{4}{c}{\textbf{FR}}   \\
    \cmidrule(lr){2-5} \cmidrule(lr){6-8} \cmidrule(lr){9-12}
    \textbf{Model} & \textbf{P} & \textbf{R} & \textbf{F1} & \textbf{Acc} & \textbf{MSE} & \textbf{MAE} & \textbf{MRE} & \textbf{P} & \textbf{R} & \textbf{F1} & \textbf{Acc} \\
    \hline
    \multirow{1}{*}{Qwen3 30B} & 0.7168 & 0.7190 & 0.7177 & 0.7086 & 81.0282 & 2.4485 & 0.061560 & 0.5760 & 0.5659 & 0.5709 & 0.5597 \\
    \hline
    \multirow{1}{*}{GPT-oss-20B} & 0.7357 & 0.7309 & 0.7315 & 0.7296 & 80.9126 & 2.4452 & 0.061472 & 0.6430 & 0.6607 & 0.6557 & 0.6229 \\
    \hline
    \multirow{1}{*}{GPT-oss-120B} & 0.7409 & 0.7423 & 0.7416 & 0.7400 & 80.2875 & 2.4093 & 0.058217 & 0.6483 & 0.6632 & 0.6652 & 0.6359 \\
    \hline
    \multirow{1}{*}{GEMMA 4-31B} & 0.7612 & 0.7534 & 0.7533 & 0.7416 & 81.3612 & 2.4652 & 0.061848 & 0.6840 & 0.6932 & 0.6875 & 0.6374 \\
    \hline
  \end{tabular}
  \caption{\label{Table 14}
    Ground Truth Workflow.}
\end{table*}

\begin{table*}
  \centering
  \small
  \begin{tabular}{llccccccc}
    \hline
    \textbf{Model} & \textbf{MSE} & \textbf{MAE} & \textbf{MRE} & \textbf{P} & \textbf{R} & \textbf{F1} & \textbf{Acc} \\ 
    \hline
    \multirow{1}{*}{Qwen3 30B} & 80.6790 & 2.4282 & 0.060553 & 0.6204 & 0.6224 & 0.6210 & 0.6029 \\
    \hline
    \multirow{1}{*}{GPT-oss-20B} & 80.8355 & 2.4346 & 0.060584 & 0.6741 & 0.6882 & 0.6712 & 0.6386 \\
    \hline
    \multirow{1}{*}{GPT-oss-120B} & 79.8980 & 2.4029 & 0.057562 & 0.6823 & 0.7043 & 0.6879 & 0.6542 \\
    \hline
    \multirow{1}{*}{GEMMA 4-31B} & 80.6177 & 2.4192 & 0.060550 & 0.6961 & 0.7080 & 0.7020 & 0.6629 \\
    \hline
  \end{tabular}
  \caption{\label{Table 15}
    Ground Truth History SQL.}
\end{table*}

\begin{table*}[hp]
  \centering
  \small
  \begin{tabular}{llccccccc}
    \hline
    \textbf{Model} & \textbf{P} & \textbf{R} & \textbf{F1} & \textbf{Acc} & \textbf{MRE} \\ 
    \hline
    \multirow{1}{*}{Qwen3 30B}   & 0.8593 & 0.8614 & 0.8603 & 0.8371 & 0.004563 \\
    \hline
    \multirow{1}{*}{GPT-oss-20B} & 0.8039 & 0.8515 & 0.8108 & 0.7971 & 0.004689 \\
    \hline
    \multirow{1}{*}{GPT-oss-120B} & 0.8186 & 0.8690 & 0.8335 & 0.8129 & 0.004153 \\
    \hline
    \multirow{1}{*}{GEMMA 4-31B} & 0.8783 & 0.8936 & 0.8859 & 0.8614 & 0.004931 \\
    \hline
  \end{tabular}
  \caption{\label{Table 16}
    Ground Truth Forecast Data.}
\end{table*}

Specifically, GT SLU yields only marginal average accuracy improvements (QR: +1.39\%; FR: +1.01\%) and the smallest MRE reduction on NF (-0.00015), indicating that the BERT-based SLU model is already strong and robust under the current setting. GT WF leads to the smallest accuracy gain (QR: +0.38\%; FR: +0.71\%) and only a limited NF MRE reduction (-0.0002); this outcome is consistent with the workflow module’s high average accuracy (99.9\%), suggesting that it is generally reliable and therefore offers relatively limited headroom for further gains.

In contrast, GT SQL produces a more noticeable improvement on FR (+0.257\%), and its NF MRE reduction (-0.000962) also exceeds that of the previous two modules, implying that SQL generation remains a component with meaningful room for optimization. This observation aligns with the SQL module’s average accuracy of approximately 99.40\%, further suggesting that it may constitute a key bottleneck in the overall pipeline.

Most notably, GT FD delivers the largest performance benefits (FR: +18.74\%) while achieving the largest MRE reduction on NF (-0.055228), highlighting substantial headroom in the temporal forecast component. Moreover, even under the GT FD setting, the highest accuracy of the forecast-based reasoning module reaches only 86.14\%, leaving a remaining gap of 13.86\% to the upper bound; this indicates that both forecast quality and reasoning over forecasts may jointly constrain the final FR performance.

\begin{figure} 
  \includegraphics[width=\linewidth]{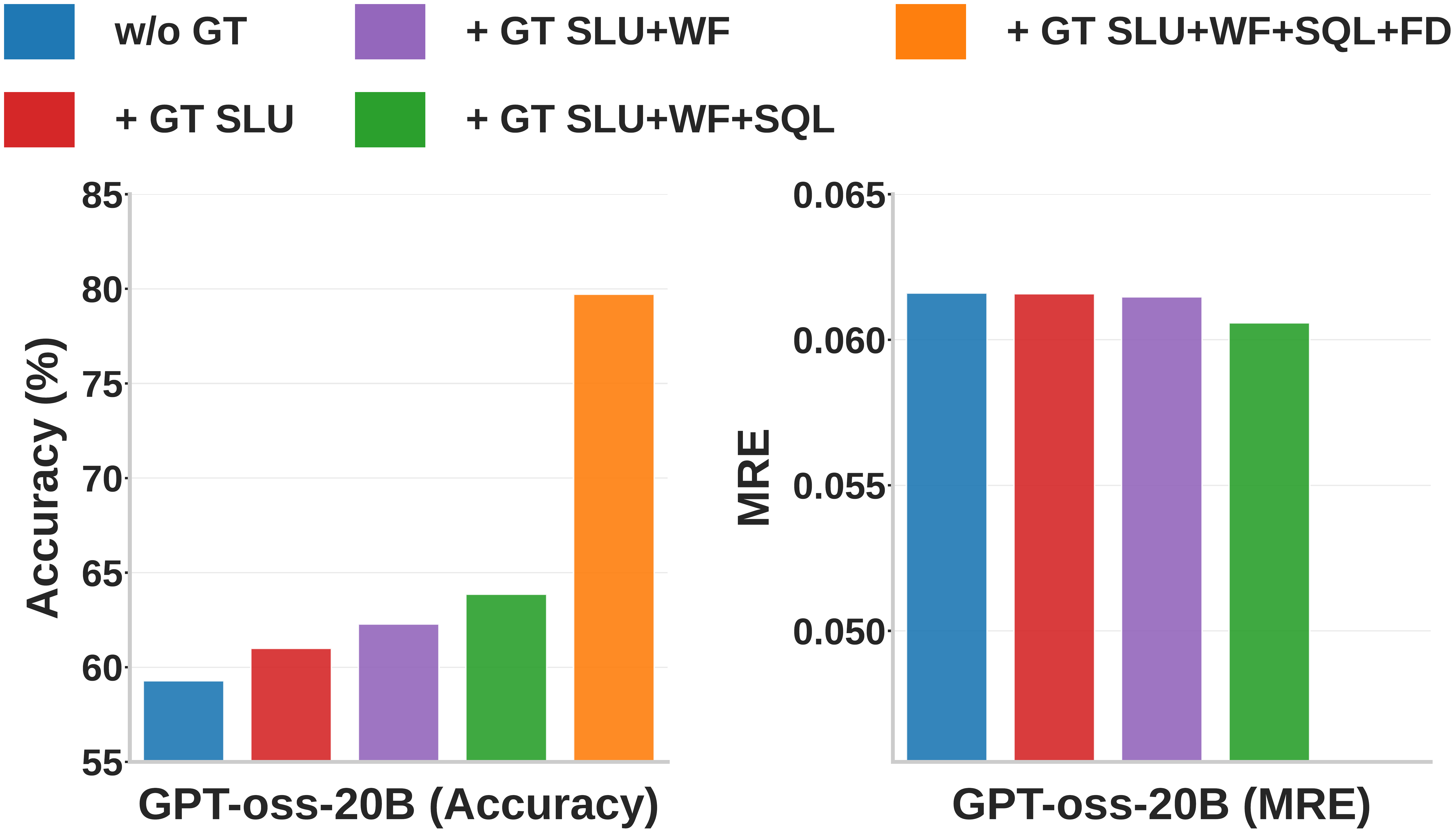}
  \caption {The impact of different GT labels on final result accuracy and MRE (GPT-oss-20B).}
  \label{fig:GPT-oss-20B_ACC_MRE}
\end{figure}

\begin{figure} 
  \includegraphics[width=\linewidth]{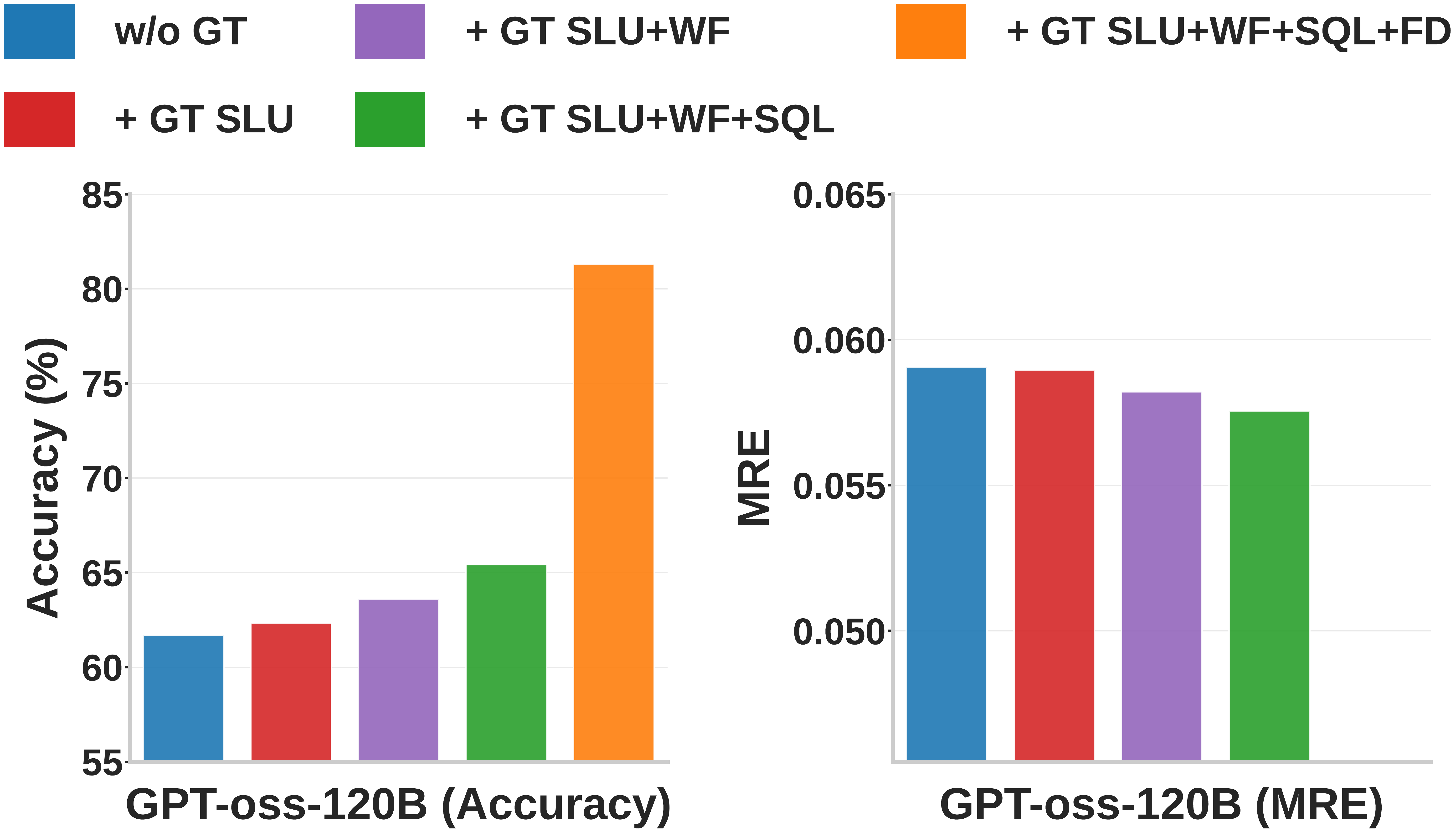}
  \caption {The impact of different GT labels on final result accuracy and MRE (GPT-oss-120B).}
  \label{fig:GPT-oss-120B_ACC_MRE}
\end{figure}

\begin{figure}
  \includegraphics[width=\linewidth]{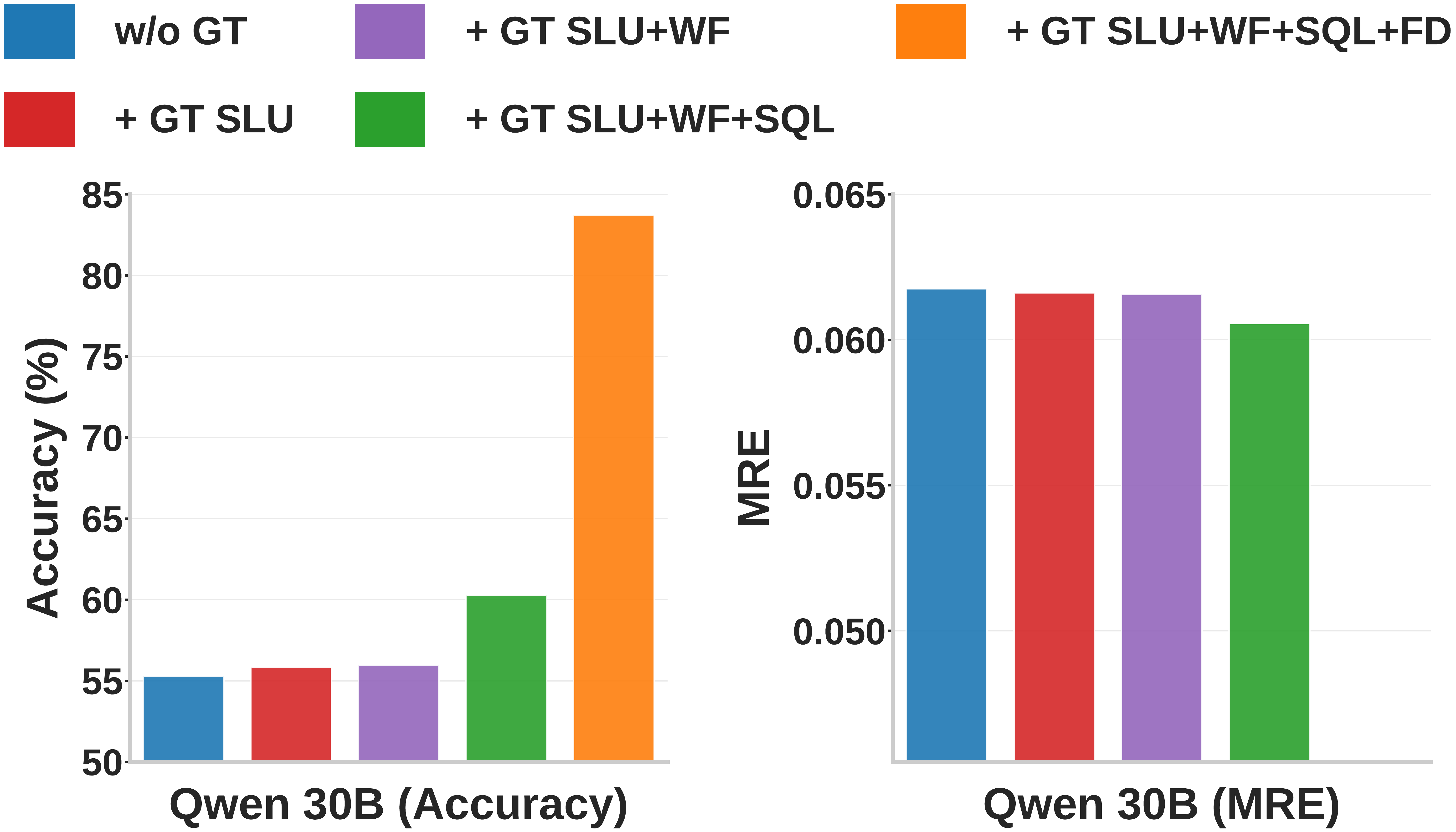}
  \caption {The impact of different GT labels on final result accuracy and MRE (Qwen3-30B).}
  \label{fig:Qwen_30B_Taller_Bars}
\end{figure}

Finally, Appendix \ref{sec:appendixD} provides visualizations of these results for Qwen3-30B (Figure \ref{fig:Qwen_30B_Taller_Bars}), GPT-oss-20B (Figure \ref{fig:GPT-oss-20B_ACC_MRE}), and GPT-oss-120B (Figure \ref{fig:GPT-oss-120B_ACC_MRE}). These figures offer an intuitive view of how different GT settings affect the performance of various backbone models.

The visualizations indicate that for forecast-based reasoning (FR) tasks, QA accuracy exhibits a sustained upward trend as different GT supervision signals are progressively incorporated. Similarly, for numerical forecasting (NF) tasks, the MRE consistently decreases with the addition of GT signals. Notably, introducing Forecast Data (FD) supervision at the final stage yields the most pronounced improvement in FR accuracy. This suggests that FD plays a pivotal role in enhancing FR performance.

However, even with all GT signals enabled, FR accuracy remains below 90\%, and the NF MRE does not converge to 0. This indicates substantial residual error and further room for improvement in both tasks. Taken together, these findings suggest that the effective acquisition and utilization of FD, as well as FR-oriented forecast-based reasoning capability, constitute the primary performance bottlenecks of the framework. This observation is consistent with the conclusions reported in the main experiments.

\section{Rewrite Result}
\label{sec:appendixE}
Appendix E reports the scoring results for questions before and after rewriting, evaluated through both large language model assessment and human evaluation for cross-validation.

\begin{figure}
  \includegraphics[width=\columnwidth]{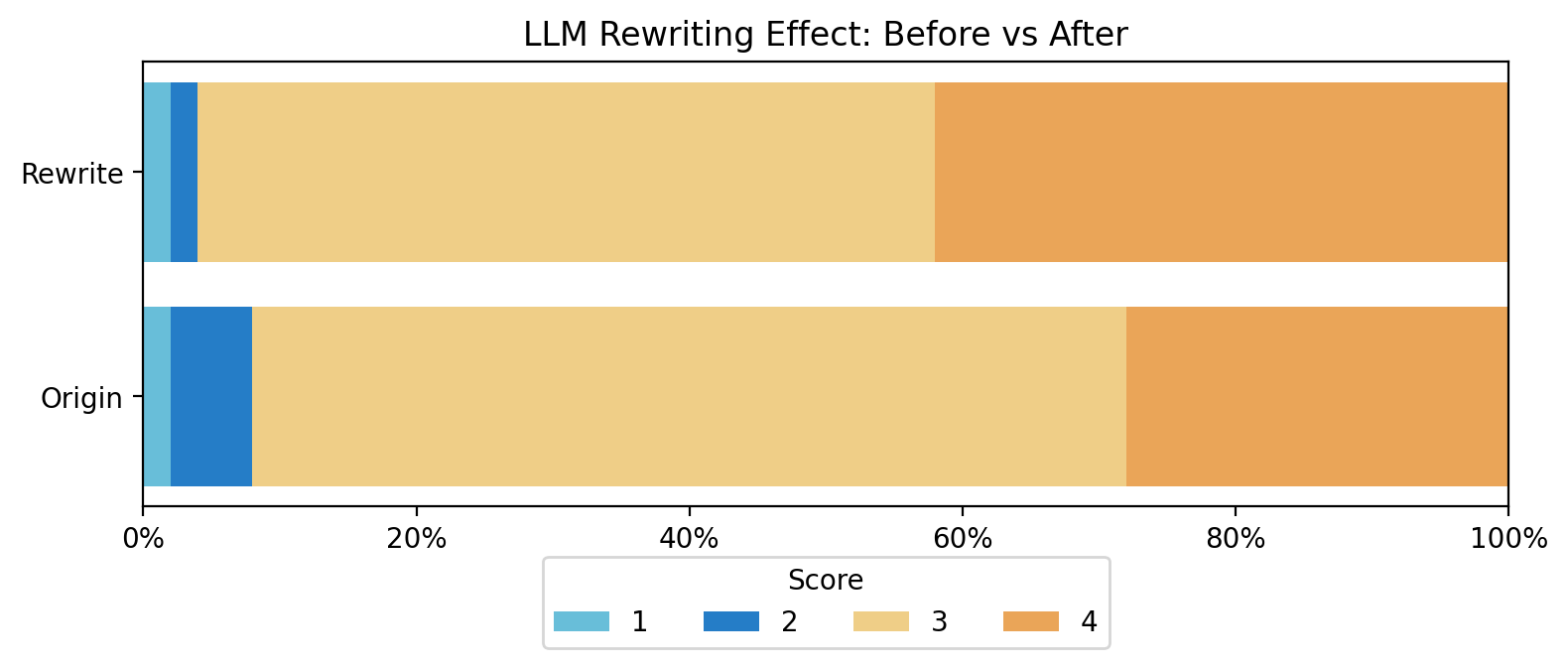}
  \caption{LLM Rewrite Effect: Before vs After.}
  \label{fig:LLM Rewrite}
\end{figure}

The LLM-based rewriting module leverages GPT-oss-120B to perform diversified paraphrasing under predefined prompt rules; while keeping the in-sentence Keywords unchanged, it enhances linguistic diversity and naturalness through transformations such as syntactic reordering (e.g., inversion) and synonym substitution (see Figure \ref{Figure 13}). To validate the effectiveness of rewriting, 50 QA pairs are randomly sampled from the pre-rewriting dataset, together with their corresponding 50 rewritten versions, resulting in a total of 100 QA instances. All instances are pre-annotated prior to scoring, then randomly shuffled for evaluation. The evaluation results are illustrated in Figure \ref{fig:LLM Rewrite}, which reports score statistics produced by the large language model Qwen3-30B. The score distribution indicates that rewritten QA pairs contain a higher proportion of high-scoring samples than the original versions, suggesting that rewriting improves overall quality under automatic evaluation criteria.

Considering that LLM evaluation may deviate from human preferences or judgment criteria, we introduce human evaluation to further verify the reliability and objectivity of the rewriting effects. Specifically, we use the same 100 QA instances (comprising 50 pre-rewriting QA pairs and their corresponding 50 rewritten versions), randomly shuffle the samples, and distribute them to 8 evaluators for independent scoring. The final score for each instance is computed as the average of the evaluators' ratings. To mitigate potential confounding factors related to task structure, all questions are pre-labeled before distribution and scoring, and are subsequently randomly assigned to evaluators.

\begin{figure} 
  \includegraphics[width=\columnwidth]{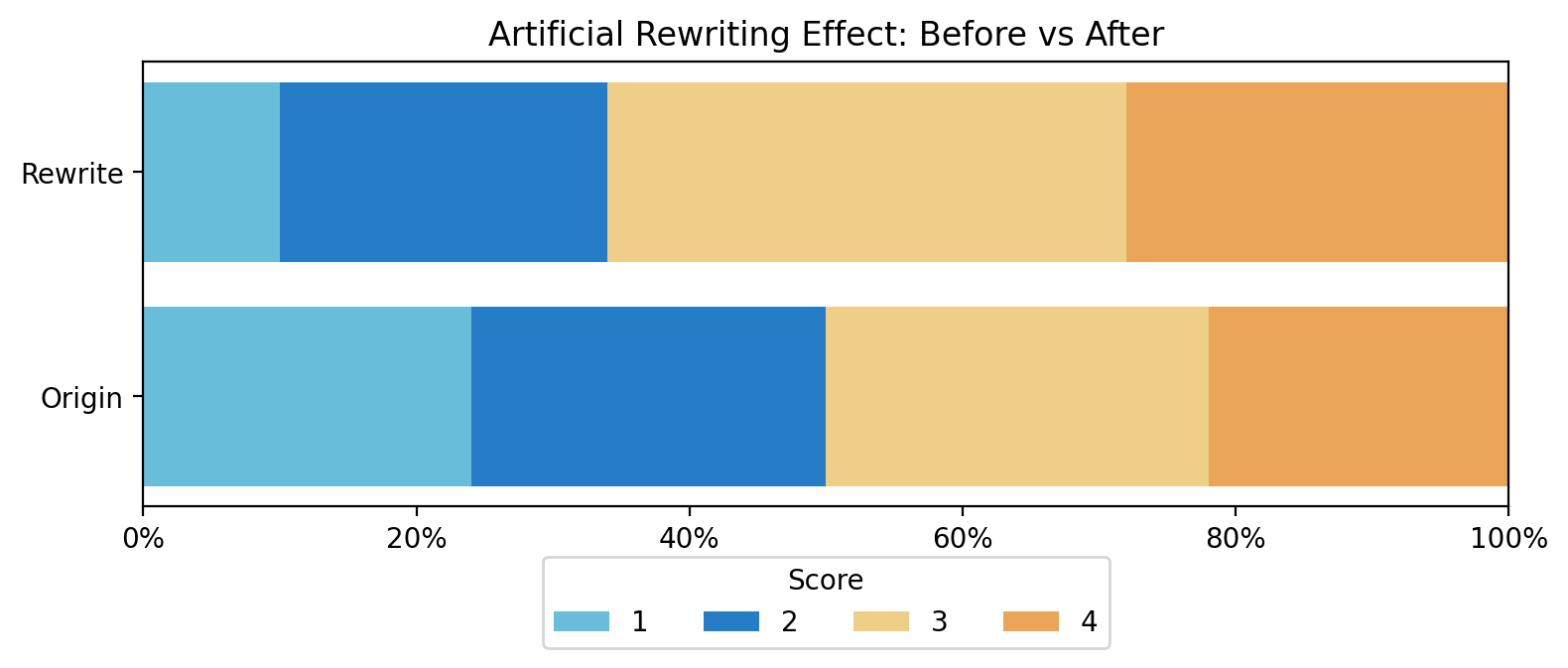}
  \caption{Artificial Rewriting Effect: Before vs After.}
  \label{fig:Artificial Rewriting}
\end{figure}

\begin{figure*}[h]
  \centering
  \includegraphics[width=0.9\textwidth]{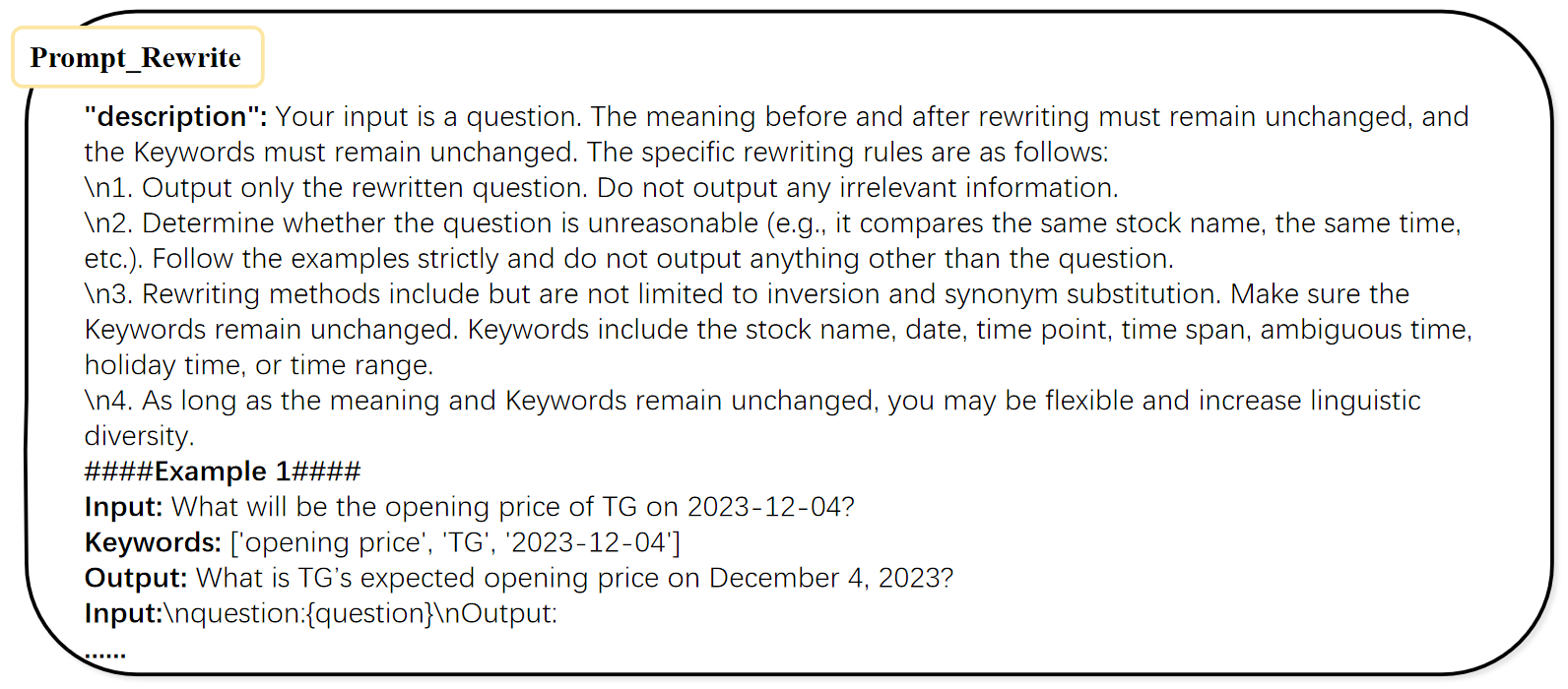}
  \caption{Prompt\_Rewriting.}
  \label{Figure 13}
\end{figure*}

\section{Prompts}
\label{sec:appendixF}
\begin{figure*}[h]
  \centering
  \includegraphics[width=0.9\textwidth]{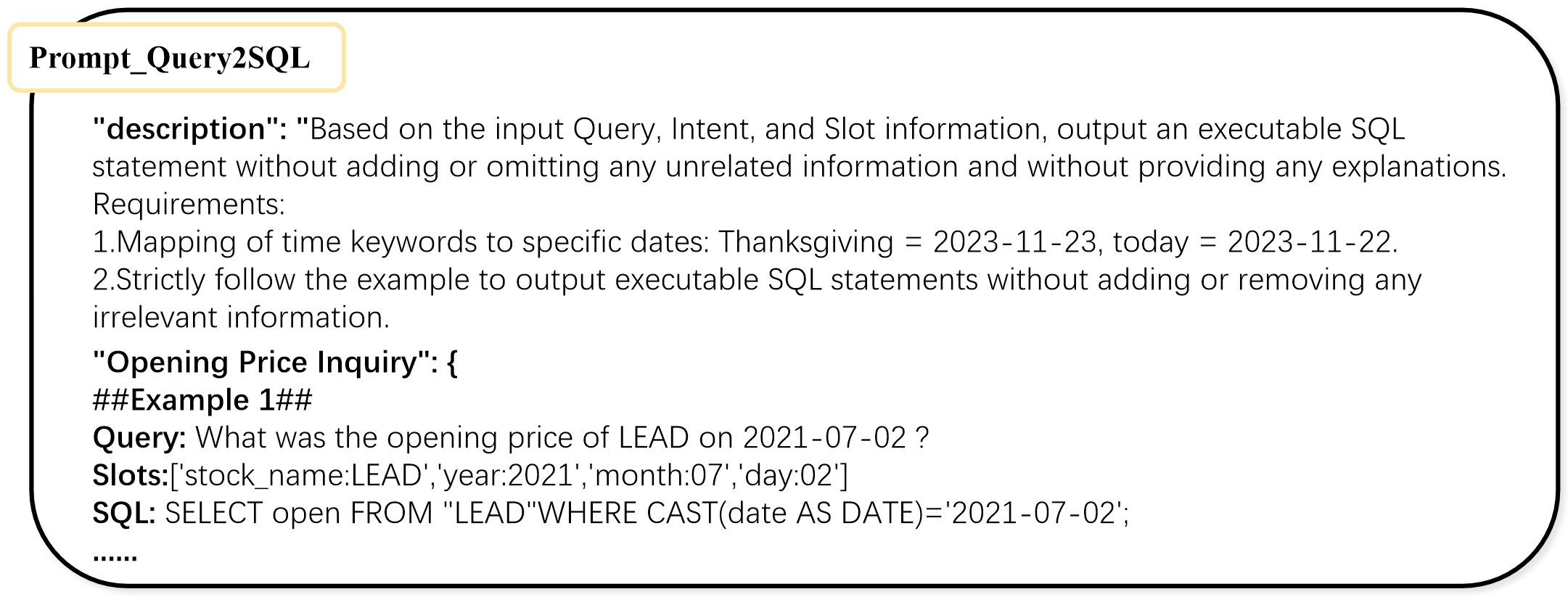}
  \caption{Prompt\_Query2SQL.}
  \label{Figure 8}
\end{figure*}
\begin{figure*}[h]
  \centering
  \includegraphics[width=0.9\textwidth]{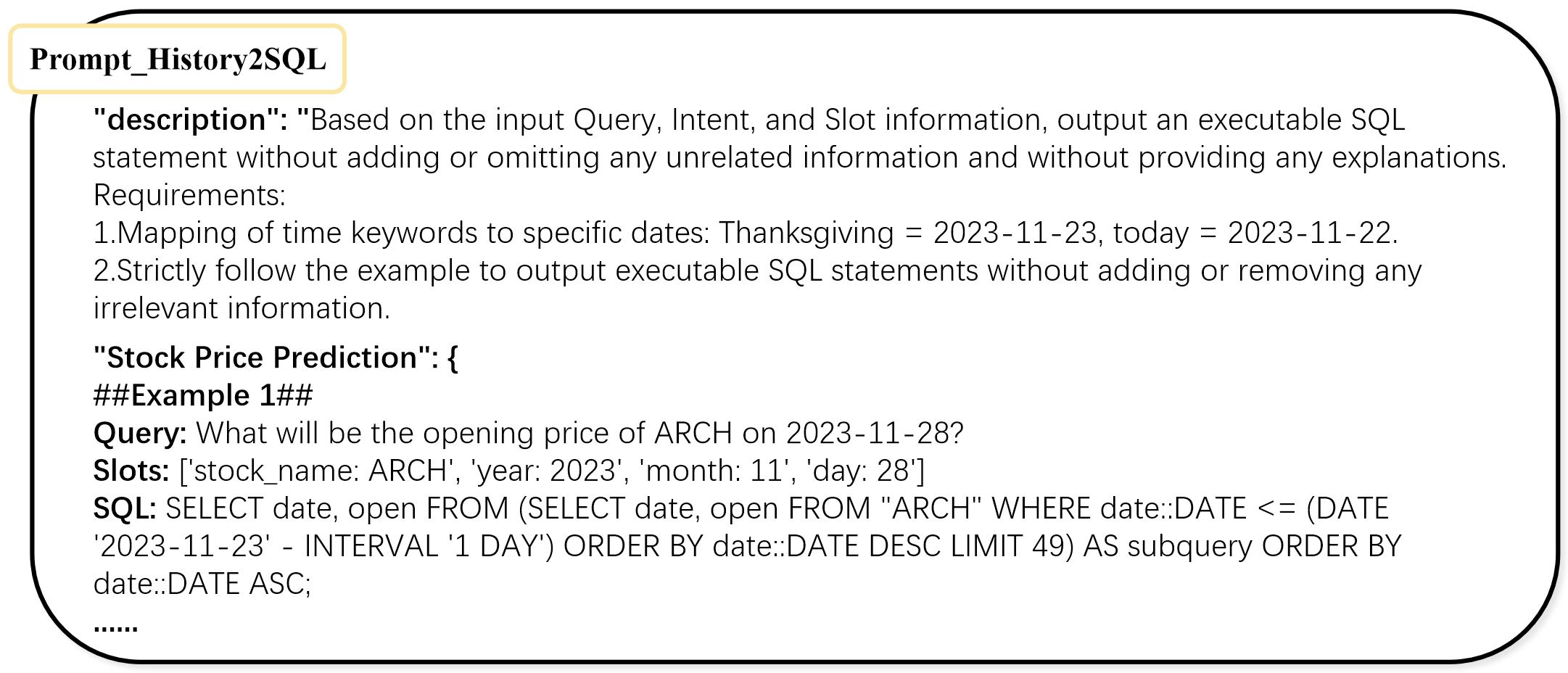}
  \caption{Prompt\_History2SQL.}
  \label{Figure 9}
\end{figure*}
\begin{figure*}[h]
  \centering
  \includegraphics[width=0.9\textwidth]{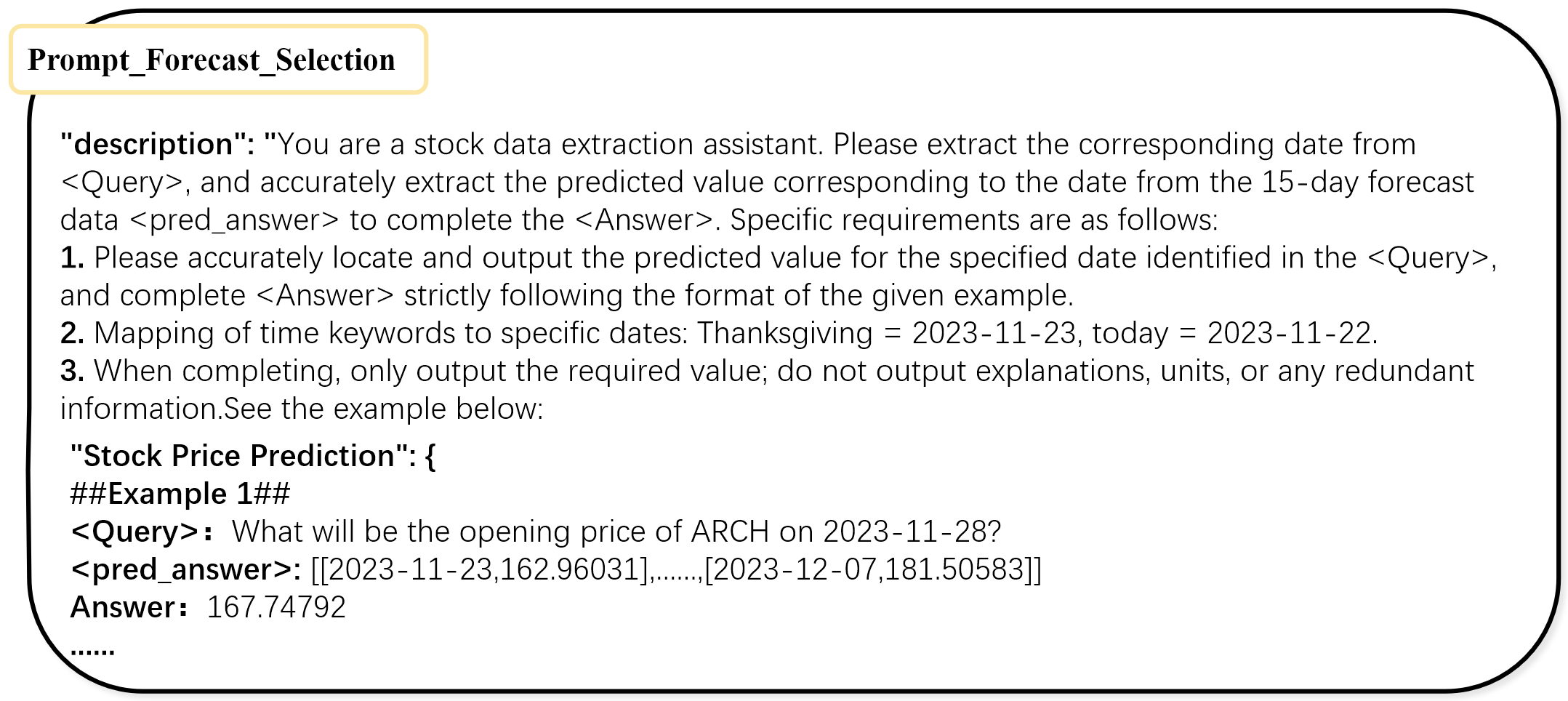}
  \caption{Prompt\_Forecast Selection.}
  \label{Figure 11}
\end{figure*}

\begin{figure*}[h]
  \centering
  \includegraphics[width=0.6\textwidth]{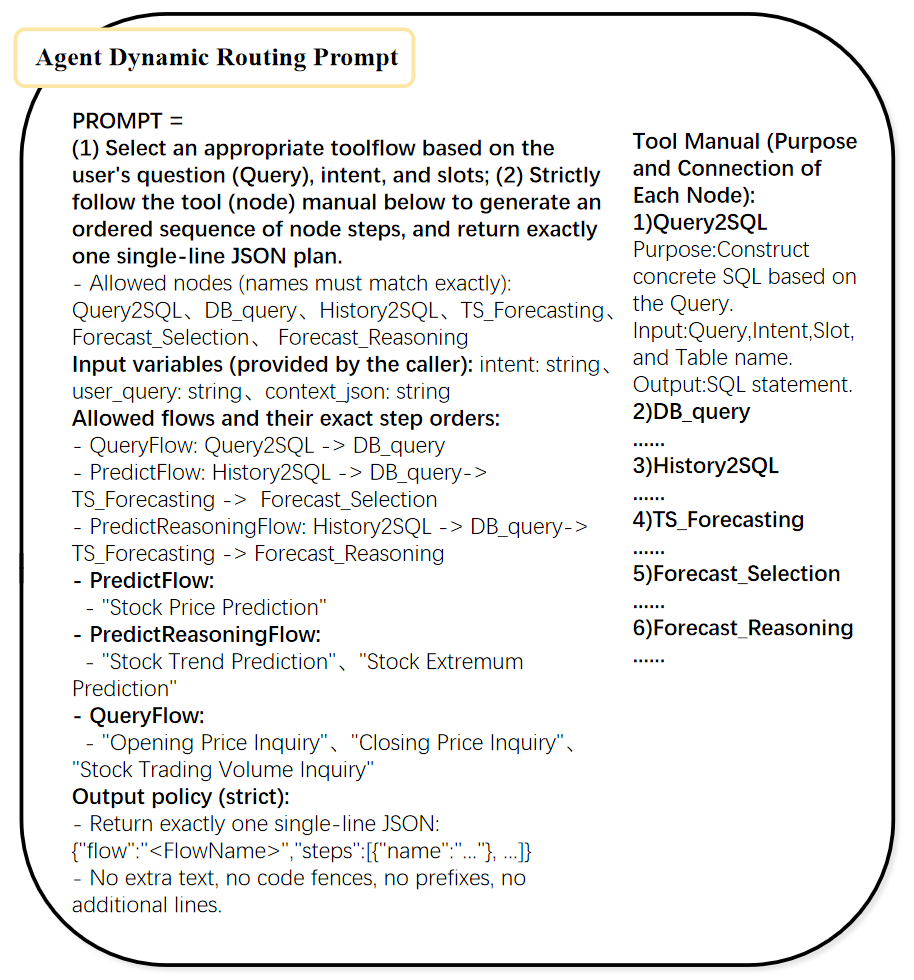}
  \caption{Prompt\_Agent Dynamic Routing.}
  \label{Figure 10}
\end{figure*} 

\begin{figure*}[h]
  \centering
  \includegraphics[width=0.9\textwidth]{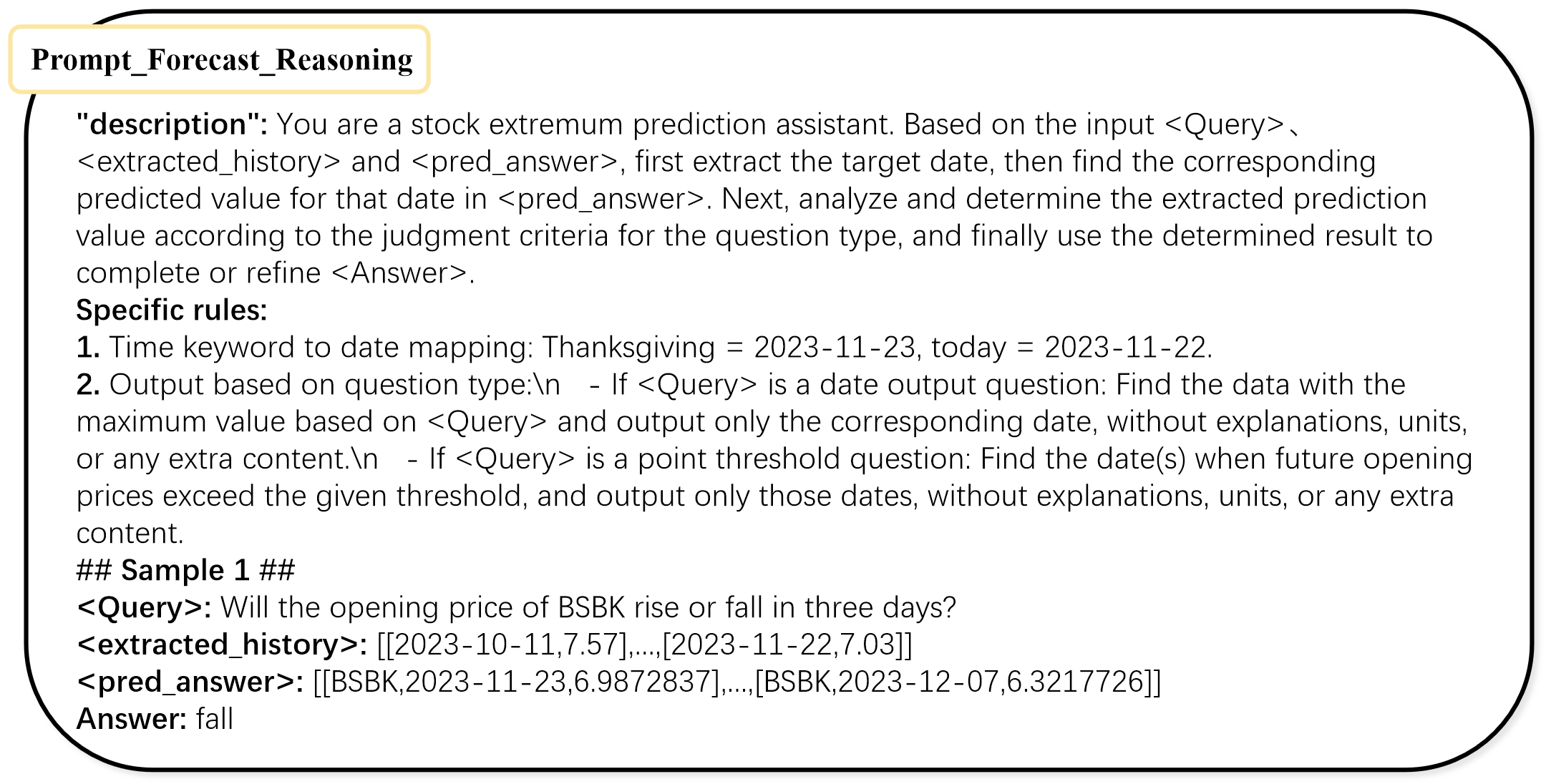}
  \caption{Prompt\_Forecast Reasoning.}
  \label{Figure 12}
\end{figure*}

Appendix F documents the prompt templates used to condition the LLM across key modules, including Prompt\_Query2SQL (Figure \ref{Figure 8}), Prompt\_History2SQL (Figure \ref{Figure 9}), Prompt\_Agent Dynamic Routing (Figure \ref{Figure 10}), Prompt\_Forecast Selection (Figure \ref{Figure 11}), and Prompt\_Forecast-based Reasoning (Figure \ref{Figure 12}). The detailed template contents are provided in the corresponding figures for reproducibility and implementation reference.

\section{Case Study}
\label{sec:appendixG}

\textit{Q: Why does the SQFRS framework still underperform even with complete GT provided?}

\begin{figure*}[tp]
  \centering
  \includegraphics[width=0.9\textwidth]{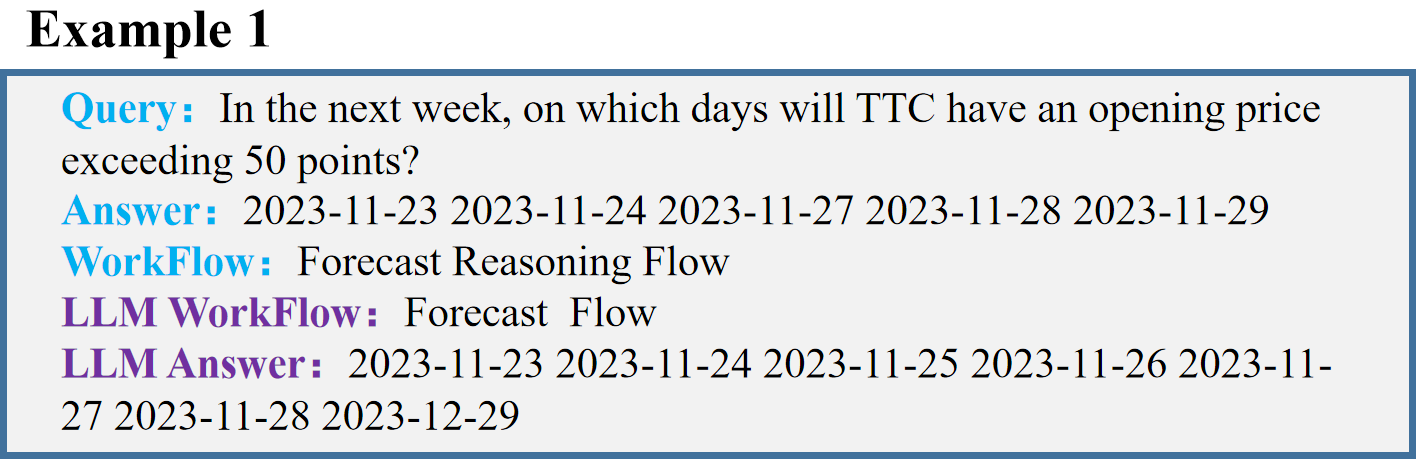}
  \caption{Example\_1.}
  \label{Figure 14}
\end{figure*}

\begin{figure*}[tp]
  \centering
  \includegraphics[width=0.9\textwidth]{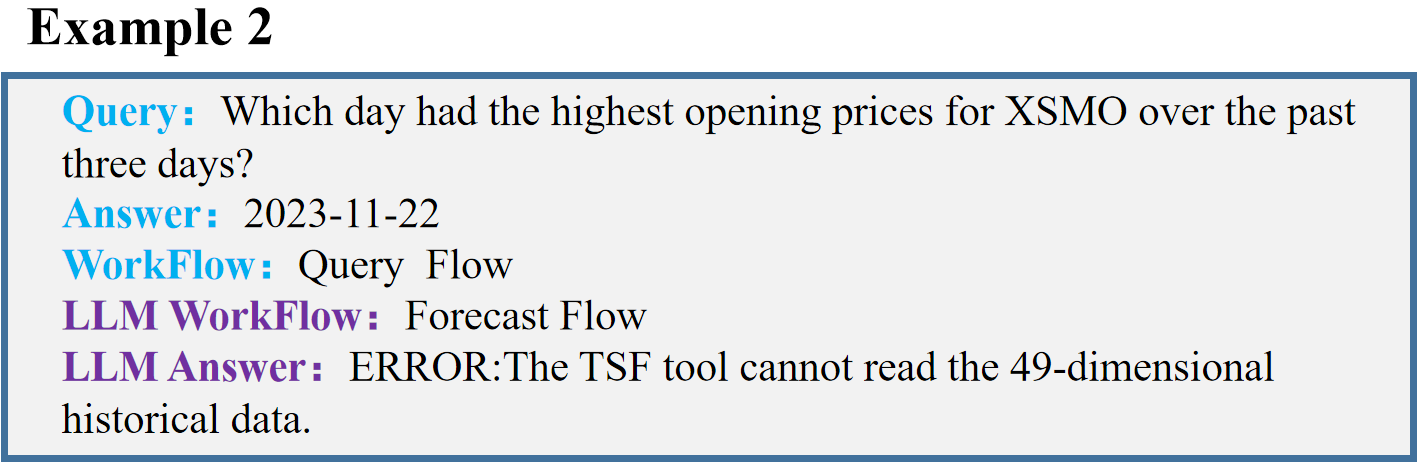}
  \caption{Example\_2.}
  \label{Figure 15}
\end{figure*}

\begin{figure*}[tp]
  \centering
  \includegraphics[width=0.9\textwidth]{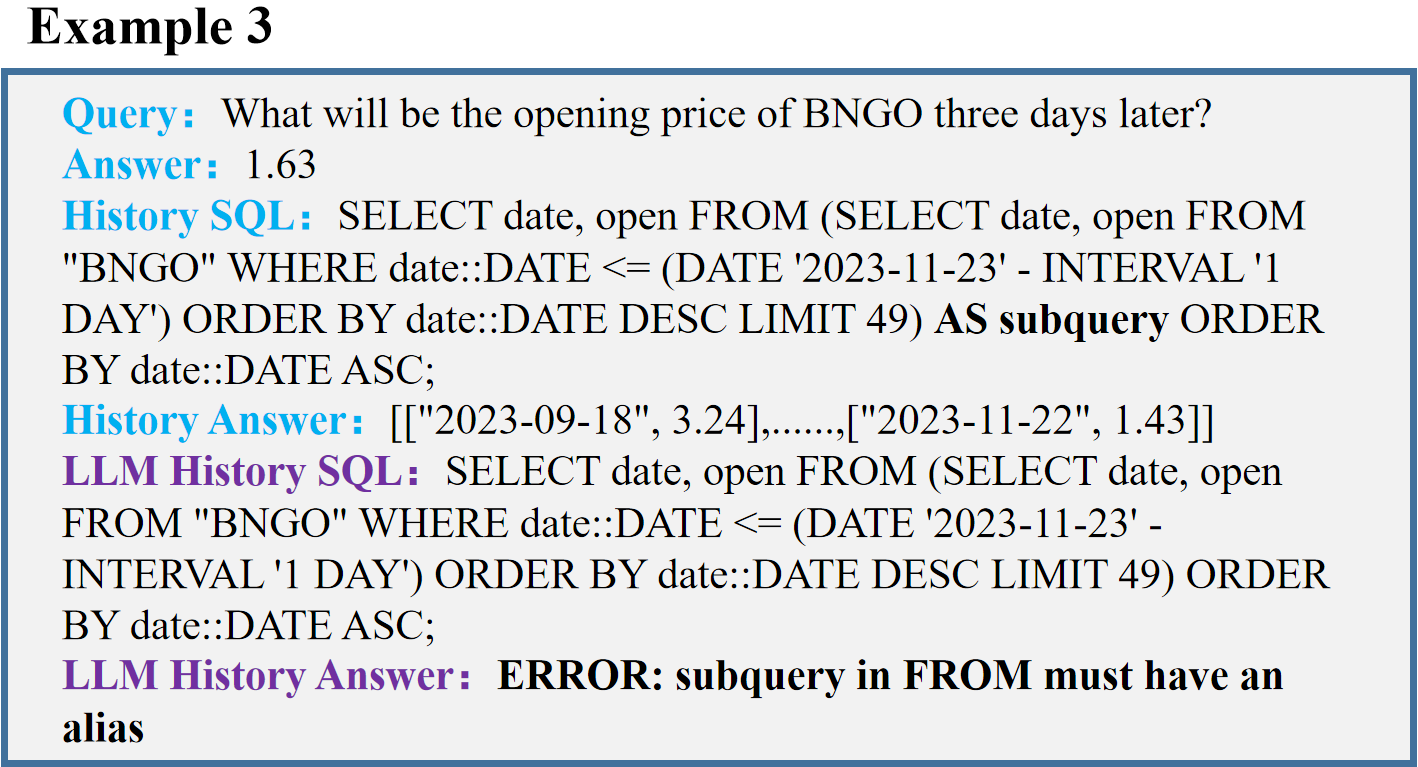}
  \caption{Example\_3.}
  \label{Figure 16}
\end{figure*}

\begin{figure*}[tp]
  \centering
  \includegraphics[width=0.9\textwidth]{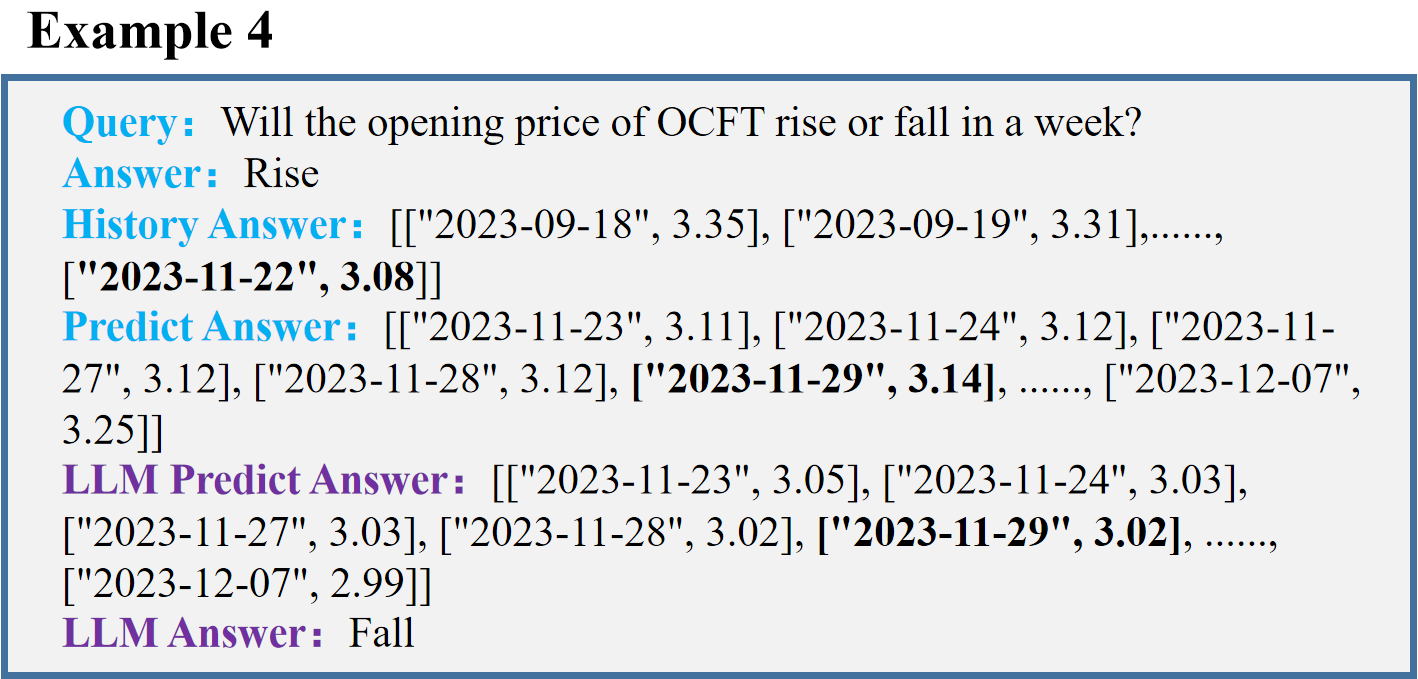}
  \caption{Example\_4.}
  \label{Figure 17}
\end{figure*}

\begin{figure*}[tp]
  \centering
  \includegraphics[width=0.9\textwidth]{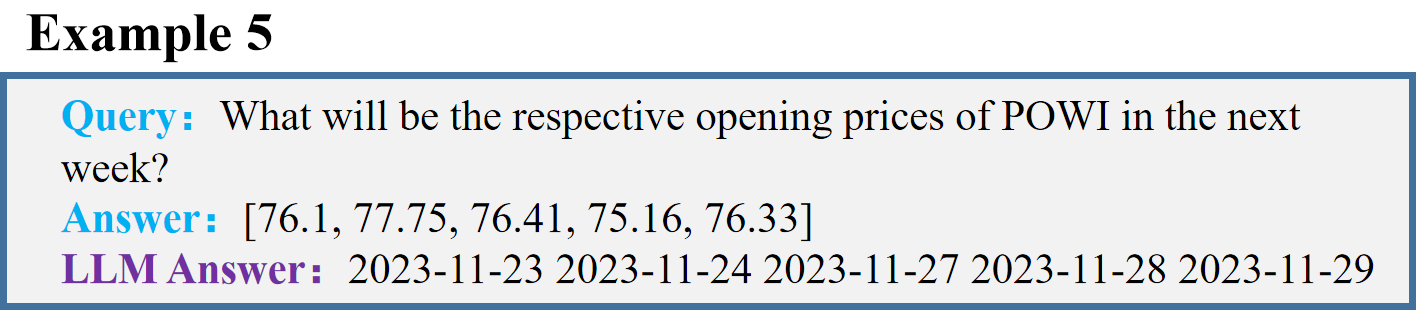}
  \caption{Example\_5.}
  \label{Figure 18}
\end{figure*}

\begin{figure*}[tp]
  \centering
  \includegraphics[width=0.9\textwidth]{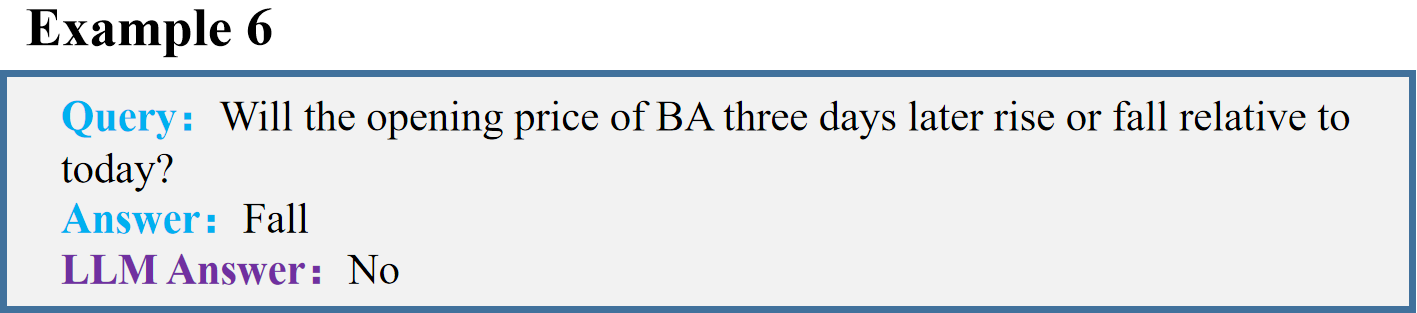}
  \caption{Example\_6.}
  \label{Figure 19}
\end{figure*}

A: To probe the capability boundary of SQFRS in tool-augmented environments, we configure the system with all available Ground Truth (GT) information. If performance remains unsatisfactory under this setting, the bottleneck likely lies in the forecasting tools and the LLM’s reasoning ability, rather than retrieval or data errors. Based on this assumption, we conduct an in-depth case study on the weakest model, Qwen3-30B, as illustrated in (Figure \ref{Figure 14}-\ref{Figure 19}). Our analysis reveals several recurrent error patterns, even when complete GT is available.

First, hallucinations caused by faulty planning. As shown in Figure \ref{Figure 14} and Figure \ref{Figure 15}, the Dynamic Routing Agent (DRA) fails to select the correct workflow, gradually deviates from the intended trajectory as it repeatedly receives negative feedback from other tools, and ultimately produces a fabricated answer. This indicates that, in multi‑agent coordination scenarios, the system remains relatively weak in task allocation and error recovery.

Second, SQL generation errors in the History2SQL module. As shown in Figure \ref{Figure 16}, even when the outputs of upstream modules such as Question→SLU are entirely correct, the History2SQL component still tends to rely on the LLM’s prior knowledge or fixed SQL templates, thereby drifting away from the current contextual query. These deviations propagate downstream along the workflow, leading to inaccurate SQL generation and preventing subsequent modules from retrieving reliable historical information.

Third, prediction inaccuracies arising from tool limitations. As shown in Figure \ref{Figure 17}, even when all modules in the Question→SLU→History2SQL→DB\_Query pipeline produce correct outputs, the Time‑Series Forecasting (TSF) module may still yield predictions with large Mean Relative Error (MRE). This suggests that, for tasks that depend on future data estimation, TSF remains one of the major limiting factors within the SQFRS framework.

Fourth, semantic hallucinations arising during the extraction of future values from TSF outputs. As illustrated in Figure \ref{Figure 18}, even when all intermediate results along the Question→SLU→History2SQL→DB\_Query→TSF workflow are correct, the Forecast Selection (FS) module can still introduce semantic errors when extracting the final result. For example, when a user asks for the “opening price,” the FS module may mistakenly extract the date instead. This indicates that, although the underlying information retrieval is accurate, the system’s understanding of answer types and semantic intent remains incomplete.

Finally, severe out of context hallucinations. As shown in Figure \ref{Figure 19}, even when modules such as Question→SLU→History2SQL→DB\_Query→TSF all produce correct outputs within the workflow, the system may still return answers that are entirely unrelated to the ground truth. Such failures suggest that, even under strong external supervision, the LLM can still generate unsupported inferences due to unstable reasoning chains, degraded decision‑making mechanisms, or misinterpretation of task objectives.


\end{document}